\documentclass[10pt,twocolumn,letterpaper]{article}

\usepackage[pagenumbers]{cvpr} 

\usepackage{amsthm}

\theoremstyle{plain}
\newtheorem{proposition}{Proposition}
\newtheorem{corollary}{Corollary}

\theoremstyle{remark}

\usepackage{algorithm}
\usepackage{algpseudocode}

\usepackage{booktabs}
\usepackage{multirow}

\usepackage{colortbl}
\definecolor{cinvbg}{gray}{0.92}

\usepackage{tabularx}

\definecolor{cvprblue}{rgb}{0.21,0.49,0.74}
\usepackage[pagebackref,breaklinks,colorlinks,allcolors=cvprblue]{hyperref}

\def\confName{CVPR}
\def\confYear{2027}

\title{Learning World Models through Contrastive Identification of Actions}

\author{Hayeong Lee\\
Korea University\\
Seoul, Republic of Korea\\
{\tt\small hayeong\_lee@korea.ac.kr}
\and
Youngjun Suh\\
Korea University\\
Seoul, Republic of Korea\\
{\tt\small youngjunsuh@korea.ac.kr}
\and
Byung-Jun Lee\\
Korea University, Gauss Labs Inc.\\
Seoul, Republic of Korea\\
{\tt\small byungjunlee@korea.ac.kr}
}

\newcommand{\method}{CIA}
\newcommand{\methodfull}{Contrastive Identification of Actions}

\begin{document}
\maketitle
\begin{abstract}
Latent world models enable goal-directed planning from offline visual data, but prediction alone does not specify which action-related distinctions their representations should preserve. Inverse dynamics provides auxiliary supervision, yet squared-error action regression rewards additional information only through changes in the conditional action mean. We propose \emph{\methodfull} (\method{}), which identifies the executed action among alternatives sampled from a history-conditioned behavior proposal. Comparing actions plausible in the same context targets information supplied by the next latent beyond history. Under exact behavior matching, we show that \method{} can reward action information that leaves optimal inverse-regression error unchanged, and verify this distinction through a controlled learning example. \method{} achieves higher planner-averaged mean success than the base world model and a regression inverse-dynamics baseline across five visual control benchmarks, without modifying the planner.
\end{abstract}
    
\section{Introduction}
\label{sec:intro}

World models support goal-directed control from reward-free offline data by predicting action consequences~\cite{ha2018world,sobal2025pldm}. Joint-embedding predictive architectures (JEPAs) learn these predictions in latent space~\cite{zhou2025dinowm,assran2025vjepa2,maes2026leworldmodel}. A planner then searches for actions whose predicted latents approach the latent of a goal image. Planning therefore succeeds only if the latent keeps the information that separates the goal from other reachable states.

Latent prediction alone does not fully specify these distinctions. Because the encoder also defines the prediction target, an observation component can be easy to predict without being useful for control. Non-collapse regularization preserves variation, but does not determine which variation matters for action selection~\cite{maes2026leworldmodel,balestriero2025lejepa}. Auxiliary objectives supply further criteria through latent trajectory geometry~\cite{wang2026straightening,fu2026control} or inverse dynamics~\cite{ivashkov2026sensorimotor,boylan2026no}. Inverse dynamics suits offline data well: every recorded transition comes with its executed action, a task-independent label that links the action to its effect.

\begin{figure}[t]
    \centering
    \includegraphics[width=\linewidth]{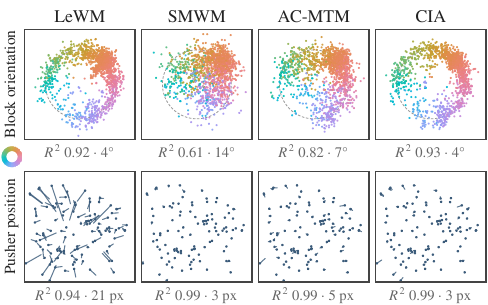}
    \caption{\textbf{Goal components decoded from frozen PushT representations.} Top: predicted block-angle coordinates; accurate readouts follow the unit circle in the true-angle color order. Bottom: shorter segments indicate better pusher localization. \method{} supports accurate readouts of both components. Numbers give test $R^2$ and median error. Further analysis is provided in \Cref{sec:failure_analysis}.}
    \label{fig:motivation}
    \vskip -0.1in
\end{figure}



Which action information should this supervision reward? Squared-error inverse regression rewards the next latent through changes in the conditional action mean. We show that additional action information can leave its minimum expected error unchanged.
For example, the next latent representation can reveal an action’s magnitude even when the conditional action mean remains unchanged.
This motivates supervision that can reward a broader range of action--outcome distinctions.

We propose \emph{\methodfull} (\method{}), an auxiliary objective for preserving action information in world-model representations.
It learns to identify the executed action among alternatives drawn from a learned history-conditioned behavior model, encouraging the next latent to retain action information beyond what the history alone provides.
Building on conditional InfoNCE~\cite{pmlr-v139-sordoni21a}, we show that, when the proposal exactly matches the history-conditioned action distribution in the offline dataset, the optimal contrastive loss is strictly below the chance-level loss whenever the next latent provides action information beyond the history, even when the conditional action mean remains unchanged.

\method{} updates the encoder shared by forward prediction and goal representation. We hypothesize that preserving distinctions among plausible action outcomes helps planners compare predicted trajectories with a goal. The proposal and score are needed only during training; planning retains the usual predictor and latent goal cost. \Cref{fig:motivation} illustrates a corresponding empirical strength: \method{} retains accurate access to both pusher position and block orientation in Push~\cite{chi2025diffusion}.


Across five visual control benchmarks and two planners, \method{} achieves higher overall mean success than LeWM and the regression-IDM baseline. These results demonstrate the practical value of \method{} as an auxiliary objective for latent world models.
Combining \method{} with CGS~\cite{fu2026control} further increases mean success on PushT and Cube~\cite{park2025ogbench}, illustrating the potential to combine contrastive action identification with geometric regularization.
Candidate-source ablations, auxiliary-network diagnostics, and goal-component analysis examine the design choices and outcomes accompanying these gains.

Our contributions are threefold:
\begin{itemize}
    \item We introduce \method{}, a behavior-informed inverse contrastive objective for world-model learning that leaves the planning procedure unchanged.
    \item We characterize the limitation of squared-error inverse regression and show how contrastive action identification with behavior-matched alternatives can reward additional action information.
    \item We demonstrate improved planning on five visual control benchmarks. Ablations of the alternative source and an analysis of planning failures show where the gains come from.
\end{itemize}



\section{Related Work}
\label{sec:related_work}

\paragraph{Representation Learning for World Models.}
JEPA-style world models support planning through latent prediction~\citep{zhou2025dinowm,assran2025vjepa2}, and LeWorldModel (LeWM) combines prediction with SIGReg~\citep{maes2026leworldmodel}. TS and CGS shape trajectory curvature and action--displacement geometry, respectively~\citep{wang2026straightening,fu2026control}. SMWM regresses the action from consecutive latents to prevent collapse~\citep{ivashkov2026sensorimotor}. Like SMWM, \method{} trains the encoder with executed actions, but it identifies them among alternatives instead of regressing them.

\paragraph{Contrastive World Models.}
C-SWM learns object-structured dynamics with contrastive state separation~\citep{kipf2020contrastive}, and Temporal Predictive Coding contrasts observed future--history pairs with unrelated trajectories~\citep{nguyen2021temporal}. AC-MTM identifies the executed action among actions from other transitions in the batch~\citep{boylan2026no}. Unlike these in-batch alternatives, \method{}'s alternatives are drawn for the same history, so they are plausible at that history.

\paragraph{Inverse Information and Control.}
GLAMOR learns inverse and action-prior models over sequences and plans with their likelihood ratio~\citep{paster2021planning}. \citet{rakelly2021sufficient} study whether information objectives yield representations sufficient for control; their inverse objective $I(A_t;Z_{t+k}\mid Z_t)$ includes our one-step target when $H=Z_t$. We use a related ratio (\Cref{eq:sstar}) only to train latents, and we ask which action information contrastive identification rewards that regression does not, rather than whether it suffices for control.

\section{Preliminaries}
\label{sec:preliminaries}

We learn an encoder and latent dynamics from an offline dataset of observation--action trajectories with no rewards or task labels~\citep{maes2026leworldmodel,zhou2025dinowm,assran2025vjepa2,wang2026straightening}.
At test time, a planner uses them to reach a goal given as an image.



For a history length $\ell$, we form an offline dataset of trajectory windows,
\begin{equation}
    \mathcal D
    =\left\{\bigl(o_{t-\ell+1:t+1},a_{t-\ell+1:t}\bigr)\right\}_{t\in\mathcal I},
    \label{eq:trajectory-windows}
\end{equation}
where $x_{i:j}=(x_i,\ldots,x_j)$, and $\mathcal I$ indexes $N$ valid windows within individual episodes. Observations satisfy $o_t\in\mathcal O$, and continuous actions satisfy $a_t\in\mathcal A\subseteq\mathbb R^{d_a}$. Here, $a_t$ denotes the control input associated with the transition from $o_t$ to $o_{t+1}$. An encoder $f_\theta:\mathcal O\to\mathcal Z\subseteq\mathbb R^d$ produces $z_t=f_\theta(o_t)$. The pre-action latent history is
\begin{equation}
    h_t=\bigl(z_{t-\ell+1:t},a_{t-\ell+1:t-1}\bigr)\in\mathcal H,
    \label{eq:latent-history}
\end{equation}
which excludes both $a_t$ and $o_{t+1}$. For $\ell=1$, the past-action sequence is empty and $h_t=z_t$. A forward model $g_\phi:\mathcal H\times\mathcal A\to\mathcal Z$ predicts $\hat z_{t+1}=g_\phi(h_t,a_t)$.

\paragraph{JEPA-based World Models.}
Following LeWM~\citep{maes2026leworldmodel}, we jointly train the encoder and predictor with one-step latent prediction:
\begin{equation}
    \mathcal{L}_{\mathrm{pred}}
    =
    \mathbb{E}_{\mathcal{D}}
    \left[
        \left\|
            g_\phi(h_t,a_t)-f_\theta(o_{t+1})
        \right\|_2^2
    \right],
    \label{eq:prediction}
\end{equation}
where the expectation is over trajectory windows sampled from the offline dataset.

To discourage collapse, we use Sketched Isotropic Gaussian Regularization (SIGReg)~\citep{balestriero2025lejepa}, which encourages latent features to follow an isotropic Gaussian distribution through their random one-dimensional projections. The base world-model objective is
\begin{equation}
    \mathcal{L}_{\mathrm{WM}}
    =
    \mathcal{L}_{\mathrm{pred}}
    +\alpha\,\mathcal{L}_{\mathrm{SIGReg}},
    \label{eq:world-model}
\end{equation}
where $\alpha>0$ controls regularization. This objective does not explicitly require identifying the executed action among alternatives plausible under the same history.


Following LeWM~\cite{maes2026leworldmodel}, we perform trajectory optimization in the learned latent space.
Given the current history and a candidate action sequence $a_{t:t+T-1}$, the predictor $g_\phi$ autoregressively rolls out future latents over a planning horizon $T$.
We optimize the action sequence to minimize the squared Euclidean distance between the terminal predicted latent and the goal embedding $z_g=f_\theta(o_g)$:
\begin{equation}
    a_{t:t+T-1}^{*}
    =
    \arg\min_{a_{t:t+T-1}}
    \left\|\hat z_{t+T}-z_g\right\|_2^2.
    \label{eq:planning-cost}
\end{equation}
The encoder and predictor remain fixed during planning.

\paragraph{Probabilistic Formulation.}
For the analysis, we drop time indices and treat a window drawn from $\mathcal D$ as random: $H$, $A$, and $Z'=f_\theta(O')$ denote its history, executed action, and next latent, and lowercase $h$, $a$, $z'$ denote their values. For a fixed encoder $f_\theta$, the data induce the joint distribution
\begin{equation}
    p(h,a,z')
    =
    p(h)\,p_\beta(a\mid h)\,p(z'\mid h,a),
    \label{eq:joint}
\end{equation}
where $p_\beta(a\mid h)$ is the \emph{behavior distribution}, namely, the conditional distribution of executed actions given the history. The term $p(z'\mid h,a)$ denotes the corresponding conditional distribution of the next latent representation.

We assume that these distributions admit densities with respect to fixed reference measures, with all integrals understood relative to those measures. The marginal distribution of the next latent is
\begin{equation}
    p(z'\mid h)
    =
    \int_{\mathcal{A}}
        p(z'\mid h,a)\,p_\beta(a\mid h)\,\mathrm{d}a.
    \label{eq:latent-marginal}
\end{equation}
By Bayes' rule, wherever $p(z'\mid h)>0$, the action posterior is
\begin{equation}
    p_\beta(a\mid h,z')
    =
    \frac{
        p(z'\mid h,a)\,p_\beta(a\mid h)
    }{
        p(z'\mid h)
    }.
    \label{eq:bayes}
\end{equation}
The posterior combines transition evidence and the behavior prior. Our analysis fixes the encoder and uses the same conditioning variable $H$ throughout.

\section{Method}
\label{sec:method}

We introduce \emph{\methodfull} (\method{}), an auxiliary objective that complements latent prediction by learning to identify the executed action. Our design combines two choices: contrasting the executed action with alternative candidates rather than regressing its value, and drawing those candidates from a proposal fitted to the history-conditioned behavior distribution.
We first explain the information rewarded by regression, then show how these choices allow \method{} to reward additional action--outcome distinctions.


\subsection{Regression Inverse Dynamics}
\label{sec:idm}

An regression IDM predicts the action from pre-action information and the next latent:
\begin{equation}
    \mathcal L_{\mathrm{IDM}}
    =\mathbb E\|A-r(H,Z')\|_2^2.
    \label{eq:idm}
\end{equation}
where $r(h,z')\in\mathbb R^{d_a}$ is the regression head. To see what this loss rewards, we compare the lowest errors reachable with three levels of input:
\begin{equation}
    \begin{aligned}
        V&:=\mathbb E\|A-\mathbb E[A]\|_2^2,\\
        V_H&:=\inf_u\mathbb E\|A-u(H)\|_2^2,\\
        V_{H,Z'}&:=\inf_r\mathbb E\|A-r(H,Z')\|_2^2.
    \end{aligned}
    \label{eq:idm-risks}
\end{equation}
$V$ is the error of the best constant guess, $\mathbb E[A]$, so it equals the total action variance. $V_H$ is the lowest error of a predictor $u(h)\in\mathbb R^{d_a}$ that sees only the history, and $V_{H,Z'}$ is the lowest error of a predictor $r(h,z')$ that also sees the next latent. The infima range over all measurable predictors with finite error, so these are the best errors that any head could reach. We assume $0<V<\infty$ and let $R_H^2:=1-V_H/V$ be the share of action variance that the history alone explains. $V_H$ and $V_{H,Z'}$ must use the same history $H$: SMWM uses $H=Z_t$~\citep{ivashkov2026sensorimotor}, and a longer history defines a different comparison.

\begin{proposition}[Contribution of the next latent]
\label{prop:idm}
The reduction in minimum error obtained by observing $Z'$ is
\begin{equation}
    \begin{aligned}
        \Delta&:=V_H-V_{H,Z'}\\
        &=\mathbb E\bigl\|\mathbb E[A\mid H,Z']
                  -\mathbb E[A\mid H]\bigr\|_2^2,
    \end{aligned}
    \label{eq:delta}
\end{equation}
with
\begin{equation}
    0\le\Delta\le V_H=(1-R_H^2)V.
    \label{eq:idm-bound}
\end{equation}
\end{proposition}

We call $\Delta$ the \emph{mean shift}: the expected change that $Z'$ causes in the conditional action mean. Squared-error regression rewards the next latent only through this mean shift. The bound adds that this reward is at most $V_H=(1-R_H^2)V$, the action variance that the history leaves unexplained; when the history already predicts the action well, the next latent can earn little under regression. The statement concerns the optimal error, not whether a trained head uses $Z'$. Appendix~\ref{app:idm} gives the proof and qualifications.

\subsection{Conditional Inverse Contrastive Objective}
\label{sec:cinv}

We identify the executed action among alternatives conditioned on the same history. Draw $(H,A_0,Z')$ from the data, with $A_0=A$, and draw $K-1$ alternatives independently from $q(\cdot\mid H)$, independently of $(A_0,Z')$ given $H$. For $K\ge2$, a shared candidate-wise score $S_\psi$ minimizes
\begin{equation}
    \mathcal L_{\mathrm{CIA}}(S_\psi;q)
    :=\mathbb E\left[
    -\log\frac{\exp S_\psi(H,Z',A_0)}
    {\sum_{j=0}^{K-1}\exp S_\psi(H,Z',A_j)}
    \right].
    \label{eq:cinv}
\end{equation}
All candidates use the same \emph{observed} next latent; no counterfactual outcomes are required.


Under the density and support assumptions in Appendix~\ref{app:cinv}, the Bayes-optimal logit is
\begin{equation}
    \begin{aligned}
        S^*(h,z',a)
        &=\log\frac{p_\beta(a\mid h,z')}{q(a\mid h)}+c(h,z')\\
        &=\log p(z'\mid h,a)
          +\log\frac{p_\beta(a\mid h)}{q(a\mid h)}
          +\widetilde c(h,z'),
    \end{aligned}
    \label{eq:sstar}
\end{equation}
where the additive terms are independent of $a$. When $q(a\mid h)=p_\beta(a\mid h)$, the behavior-prior log-ratio vanishes. The optimal score then ranks candidate actions by the likelihood $p(z'\mid h,a)$ of the observed next latent given the history and each candidate action.
This motivates fitting a history-conditioned proposal to offline actions. For a fitted proposal, the same expression describes compatibility relative to $q$, including any remaining prior-ratio contribution. This construction builds on contrastive estimation and conditional InfoNCE~\citep{oord2018cpc,ma-collins-2018-noise,pmlr-v139-sordoni21a}.

\begin{algorithm}[t!]
\caption{Contrastive Identification of Actions}
\label{alg:cinv}
\begin{algorithmic}[1]
\Require Offline trajectories $\mathcal D$, candidate count $K$, weight $\lambda$
\State Initialize $f_\theta$, $g_\phi$, $S_\psi$, and $q_\eta$
\State Prefit $\eta$ using \eqref{eq:cinv-proposal}, with the world model frozen
\For{each training iteration}
  \State Sample trajectory windows and their executed actions $a$ from $\mathcal D$
  \State Construct $h$ and observed next latents $z'=f_\theta(o')$
  \State Set $a_0=a$; draw and detach $K-1$ alternatives from the pre-update $q_\eta(\cdot\mid\operatorname{sg}(h))$
  \State Update $\eta$ only using \eqref{eq:cinv-proposal}
  \State Jointly permute the stored candidates and positive label
  \State Compute \eqref{eq:cinv} using the same $z'$ for every candidate
  \State Update $(\theta,\phi,\psi)$ using \eqref{eq:cinv-total}, holding $q$ and candidates fixed
\EndFor
\State \Return $f_\theta,g_\phi$ for planning
\end{algorithmic}
\end{algorithm}

\begin{proposition}[Action identification beyond history]
\label{prop:cinv-matched}
If $q(\cdot\mid H)=p_\beta(\cdot\mid H)$ almost surely, every score $s(H,A)$ ignoring $Z'$ satisfies
\begin{equation}
    \mathcal L_{\mathrm{CIA}}(s;p_\beta)\ge\log K.
    \label{eq:cinv-history-floor}
\end{equation}
For every score $S$,
\begin{equation}
    I(A;Z'\mid H)\ge\log K-\mathcal L_{\mathrm{CIA}}(S;p_\beta).
    \label{eq:cinv-cmi}
\end{equation}
\end{proposition}

Under matching, improvement over chance requires information from the next latent beyond $H$. The following result makes the contrast with regression explicit.


\begin{corollary}[Action information beyond conditional means]
\label{cor:cinv-dependence}
Under the assumptions of \Cref{prop:idm,prop:cinv-matched}, fix the encoder and $H$. If
\[
    \mathbb E[A\mid H,Z']=\mathbb E[A\mid H]
    \quad\text{almost surely},
\]
but $I(A;Z'\mid H)>0$, then
\begin{equation}
    \begin{aligned}
        V_H-V_{H,Z'}=0, \quad
        \inf_S\mathcal L_{\mathrm{CIA}}(S;p_\beta)<\log K,
    \end{aligned}
    \label{eq:mean-preserving-contrast}
\end{equation}
where $K\ge2$ and scores range over unrestricted measurable candidate-wise functions.
\end{corollary}

\Cref{cor:cinv-dependence} gives \method{} a broader criterion for preserving action information: changes in the conditional action distribution can lower the optimal contrastive loss even when its mean remains unchanged. The proof is in Appendix~\ref{app:cinv-corollary-proof}. We illustrate the same distinction below by adding distance information to a representation that already preserves direction.

\begin{figure}
    \centering
    \includegraphics[width=\linewidth]{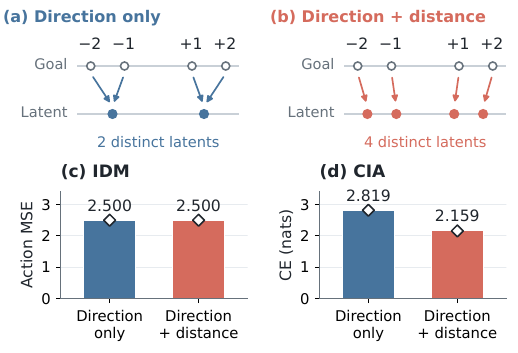}
    \caption{\textbf{Rewarding information beyond the conditional action mean.} Top: Fixed encodings preserve direction alone or both direction and distance. Bottom: Adding distance leaves learned IDM MSE unchanged but reduces \method{} loss. Bars show mean test losses across five initializations; diamonds mark theoretical optima.}
    \label{fig:cinv_information_reward}
    \vskip -0.1in
\end{figure}


\begin{table*}[t]
\centering
\caption{\textbf{Goal-conditioned planning success rates (\%).} We report the mean and standard error across five training seeds, using 150 test tasks per seed and planner. The last row reports, for each method, the unweighted mean over all ten environment--planner pairs. The highest and second-highest method means in each row are shown in bold and underlined, respectively.}
\label{tab:planning_mppi_cem}
\small
\setlength{\tabcolsep}{4pt}
\renewcommand{\arraystretch}{1.12}
\newcolumntype{Y}{>{\centering\arraybackslash}X}
\begin{tabularx}{\textwidth}{
    llYYYYY
    >{\columncolor{cinvbg}[\tabcolsep][2pt]}Y
    >{\columncolor{cinvbg}[2pt][\tabcolsep]}Y}
\toprule
Environment & Planner & LeWM & TS & CGS & SMWM & AC-MTM
& \method{} & \method{}+CGS \\
\midrule

\multirow{2}{*}{PushT}
& MPPI
& $65.3 \pm 1.5$
& $59.5 \pm 2.8$
& $71.3 \pm 1.3$
& $76.9 \pm 1.2$
& $80.0 \pm 0.8$
& $\underline{81.9} \pm 1.1$
& $\mathbf{82.1} \pm 1.0$ \\
& CEM
& $88.9 \pm 0.7$
& $85.7 \pm 1.4$
& $89.5 \pm 0.6$
& $88.3 \pm 1.5$
& $89.2 \pm 1.1$
& $\underline{91.1} \pm 0.5$
& $\mathbf{92.9} \pm 0.4$ \\
\midrule

\multirow{2}{*}{Reacher}
& MPPI
& $59.6 \pm 0.7$
& $62.5 \pm 1.2$
& $63.5 \pm 0.4$
& $65.2 \pm 0.6$
& $63.6 \pm 1.3$
& $\underline{67.3} \pm 1.0$
& $\mathbf{68.9} \pm 2.5$ \\
& CEM
& $78.8 \pm 0.1$
& $77.2 \pm 0.7$
& $78.8 \pm 1.3$
& $80.4 \pm 1.2$
& $\underline{80.5} \pm 0.8$
& $\mathbf{82.0} \pm 0.8$
& $76.8 \pm 1.9$ \\
\midrule

\multirow{2}{*}{Cube}
& MPPI
& $61.7 \pm 0.5$
& $70.3 \pm 0.9$
& $\underline{75.9} \pm 0.6$
& $68.1 \pm 0.7$
& $74.5 \pm 0.2$
& $74.7 \pm 2.3$
& $\mathbf{78.9} \pm 0.5$ \\
& CEM
& $63.6 \pm 0.3$
& $67.1 \pm 1.1$
& $\underline{77.2} \pm 0.3$
& $69.2 \pm 0.4$
& $73.2 \pm 1.2$
& $72.5 \pm 2.7$
& $\mathbf{78.8} \pm 0.9$ \\
\midrule

\multirow{2}{*}{Finger}
& MPPI
& $14.8 \pm 1.0$
& $15.1 \pm 0.8$
& $13.2 \pm 0.4$
& $\mathbf{19.1} \pm 0.9$
& $17.3 \pm 1.0$
& $16.7 \pm 0.9$
& $\underline{18.0} \pm 1.0$ \\
& CEM
& $28.4 \pm 0.7$
& $31.3 \pm 2.1$
& $30.1 \pm 0.1$
& $27.7 \pm 2.0$
& $26.9 \pm 2.0$
& $\mathbf{43.2} \pm 1.1$
& $\underline{40.0} \pm 1.6$ \\
\midrule

\multirow{2}{*}{Ball-in-Cup}
& MPPI
& $16.1 \pm 0.6$
& $14.5 \pm 0.9$
& $17.7 \pm 1.3$
& $15.3 \pm 0.8$
& $12.7 \pm 0.8$
& $\mathbf{20.7} \pm 1.6$
& $\underline{19.6} \pm 1.3$ \\
& CEM
& $38.1 \pm 1.6$
& $38.4 \pm 3.4$
& $34.7 \pm 2.3$
& $29.2 \pm 1.6$
& $35.1 \pm 2.0$
& $\mathbf{58.0} \pm 2.0$
& $\underline{54.1} \pm 2.6$ \\
\midrule

\multicolumn{2}{l}{Overall}
& $51.5$
& $52.2$
& $55.2$
& $53.9$
& $55.3$
& $\underline{60.8}$
& $\mathbf{61.0}$ \\
\bottomrule
\end{tabularx}
\end{table*}

\paragraph{Direction--Distance Example.}
We illustrate this contrast with two fixed encodings. From a fixed start state, the action is $A=(D,B)$. $D$ is uniform on $\{-1,1\}$, and $B$ is independently uniform on $\{-2,-1,1,2\}$. The agent ends at $X'=D|B|$, so $D$ sets the direction and $|B|$ sets the distance. One encoding keeps direction only, and the other keeps direction and distance; both have unit variance. Given direction, distance reveals $|B|$ but not the sign of $B$, so it changes the distribution of $B$ without moving its mean. Both encodings therefore give the same conditional action mean, $(D,0)$, and the same optimal MSE, $5/2$. Keeping distance earns nothing under squared-error regression, even though it separates near goals from far ones. This is the pattern of \Cref{cor:cinv-dependence}, applied between two encodings rather than between $H$ and $(H,Z')$ (Appendix~\ref{app:cinv-control-example}).

We train IDM heads and \method{} scores on each encoding, drawing the 32 alternatives from the uniform behavior policy, which is the matched proposal here. Adding distance leaves the learned IDM MSE unchanged but lowers the \method{} cross-entropy, close to the theoretical optima of about $2.82$ and $2.16$ nats (\Cref{fig:cinv_information_reward}). With distance, the score can reject alternatives that have the right direction but the wrong magnitude. \method{} therefore rewards the information that separates near and far goals, although regression does not. Appendix~\ref{app:direction_distance_example} gives the derivations, training details, and a separate joint-learning check.


\paragraph{Learning the Proposal.}
For a proposal, the largest identification gain available without the next latent is
\begin{equation}
    b(q):=\log K-\inf_{s(H,A)}
    \mathcal L_{\mathrm{CIA}}(s;q).
    \label{eq:cinv-bias-main}
\end{equation}
The corrected bound is
\begin{equation}
    I(A;Z'\mid H)
    \ge \log K-\mathcal L_{\mathrm{CIA}}(S;q)-b(q).
    \label{eq:cinv-mismatch-main}
\end{equation}
The correction $b(q)$ accounts for proposal-induced identification without $Z'$. To fit history-conditioned alternatives to the behavior, we minimize
\begin{equation}
    \mathcal L_q(\eta)
    :=-\mathbb E_{H,A}
    \log q_\eta\bigl(A\mid\operatorname{sg}(H)\bigr).
    \label{eq:cinv-proposal}
\end{equation}
For fixed history features, minimizing this objective minimizes the expected forward KL from the behavior within the proposal family. This KL upper-bounds $b(q)$ (Appendix~\ref{app:cinv-mismatch}), giving a principled fitting objective without requiring exact matching for the training procedure. We use a separate optimizer and detach sampled alternatives.

\paragraph{World-model Training.}
We retain the prediction loss and latent regularization:
\begin{equation}
    \mathcal L_{\mathrm{train}}
    =\mathcal L_{\mathrm{WM}}+\lambda\mathcal L_{\mathrm{CIA}}.
    \label{eq:cinv-total}
\end{equation}

\method{} directly updates the encoder and score; the predictor learns through the prediction loss.
\Cref{alg:cinv} summarizes the updates; implementation details are in Appendix~\ref{app:cinv-implementation}.
\section{Experiments}
\label{sec:experiments}

We first evaluate goal-conditioned planning performance and the quality of the learned proposal and score. We then analyze which goal components account for the gains and examine the contribution of alternative-action sources through a controlled ablation.

\subsection{Planning Evaluation Setup}
\label{sec:planning_setup}

We evaluate five image-based control environments with continuous actions: PushT~\cite{chi2025diffusion}, OGBench-Cube~\cite{park2025ogbench}, and Reacher, Finger (\texttt{turn\_easy}), and Ball-in-Cup (\texttt{catch}) from the DeepMind Control Suite~\cite{tassa2018deepmind}. Tasks require reaching a visual goal configuration; environment, dataset, and success definitions are in Appendix~\ref{app:env_dataset}.

We compare LeWM~\cite{maes2026leworldmodel}, Temporal Straightening (TS)~\cite{wang2026straightening}, Control-Geometry Straightening (CGS)~\cite{fu2026cgs}, SMWM-style inverse regression~\cite{ivashkov2026sensorimotor}, AC-MTM contrastive inverse learning~\cite{boylan2026no}, and \method{}.
All methods use the LeWM backbone.
Although their original formulations do not require SIGReg, we retain it in our LeWM-based adaptations of AC-MTM and SMWM to compare auxiliary objectives under the same non-collapse regularization.
We also evaluate \method{}+CGS to examine how contrastive action identification and geometric regularization work together.
\method{} uses a learned history-conditioned proposal $q_\eta(a\mid h)$ in PushT, Cube, Finger, and Ball-in-Cup, and a fixed-uniform proposal matching the known uniform behavior distribution in Reacher. Training preprocessing and configurations are detailed in Appendix~\ref{app:env_dataset} and \ref{app:hyperparams}.

MPPI~\cite{williams2017mppi} and CEM~\cite{deboer2005tutorial} search over predicted latent goal costs with the same evaluation protocol within each environment: execute a 25-primitive-step plan, then replan once, for at most 50 steps.
Each method uses five training seeds and the same 150 test tasks per seed and planner. \Cref{tab:planning_mppi_cem} reports means and standard errors across seeds.

\subsection{Goal-Conditioned Planning Performance}
\label{sec:planning_results}

As shown in \Cref{tab:planning_mppi_cem}, \method{} delivers strong planning performance on its own, achieving higher mean success than LeWM in every environment--planner combination with an aggregate gain of $9.3\%p$ under equal weighting of environments and planners. These results support adding action-discrimination supervision to latent prediction and non-collapse regularization. The gains occur with both MPPI and CEM, without changing the planning algorithm or latent goal cost.

The comparison with SMWM highlights the practical value of contrastive action identification: \method{} achieves higher overall planner-averaged mean success, with the largest gains on Finger and Ball-in-Cup.
Under exact behavior matching, this objective can reward distinctions beyond the conditional action mean (\Cref{cor:cinv-dependence}). When retained by the shared encoder, these distinctions can inform forward prediction and goal comparison, providing a possible explanation for the planning gains examined by the following diagnostics.
To control for the inverse input, we also evaluate SMWM-H with \method{}'s history and query construction; \method{} also retains a higher overall mean than SMWM-H, which matches its inverse input and query construction. The clearest gains occur on Ball-in-Cup and Finger; environment-specific results and bootstrap intervals are reported in Appendix~\ref{app:smwm_history}.

AC-MTM provides a comparison, as it also trains representations through contrastive inverse-action identification.
\method{} achieves higher planner-averaged means in four environments, with the clearest advantages on Finger and Ball-in-Cup.
Whereas AC-MTM uses in-batch alternatives, \method{} draws candidate actions conditioned on the same history. This construction aims to emphasize action--outcome compatibility within the same behavioral context. The methods also differ in score parameterization and candidate count; \Cref{sec:ablation} isolates candidate-source effects within \method{}.

On Cube, \method{} achieves competitive performance, and increasing the MPPI sampling budget narrows its gap to CGS.
Diagnostics of the same frozen models suggest that geometric regularization could further improve planning when combined with action identification (Appendix~\ref{app:cube_planning_geometry}).
\method{}+CGS achieves the highest planner-averaged means on Cube and PushT and a slightly higher overall mean than \method{} alone ($61.0\%$ versus $60.8\%$). Standalone \method{} retains higher means on Reacher, Finger, and Ball-in-Cup.
The combined configurations use separately specified weights,
so these results demonstrate the potential of the combination rather than an isolated additive effect or a uniform benefit across environments.

\begin{table}[t]
  \centering
  \caption{\textbf{Proposal and score quality.} Validation means over three seeds. Boundary rates are shown as $q$/data; Other $z'$ uses the next latent from another query. CE is in nats.}
\label{tab:proposal_score}
  \small
  \setlength{\tabcolsep}{3pt}
  \begin{tabular}{@{}lccccc@{}}
    \toprule
                & \multicolumn{2}{c}{Proposal $q_\eta$} & \multicolumn{3}{c}{Score $S_\psi$} \\
    \cmidrule(lr){2-3}\cmidrule(lr){4-6}
    Environment & \shortstack{Sample\\var.} & \shortstack{Boundary\\(\%)} & \shortstack{Acc.\\(\%)} & CE & \shortstack{Other $z'$\\acc. (\%)} \\
    \midrule
    PushT       & 0.19 & 0.03 / 0.02 & 57.6 & 1.49 & 4.3 \\
    Cube        & 0.24 & 7.2 / 7.4   & 89.0 & 0.34 & 3.9 \\
    Ball-in-Cup & 0.25 & 0.8 / 0.6   & 89.7 & 0.27 & 2.9 \\
    Finger      & 0.61 & 11.2 / 10.9 & 80.1 & 0.51 & 2.3 \\
    \bottomrule
  \end{tabular}
\end{table}

\begin{figure*}[t]
    \centering
    \includegraphics[width=\linewidth]{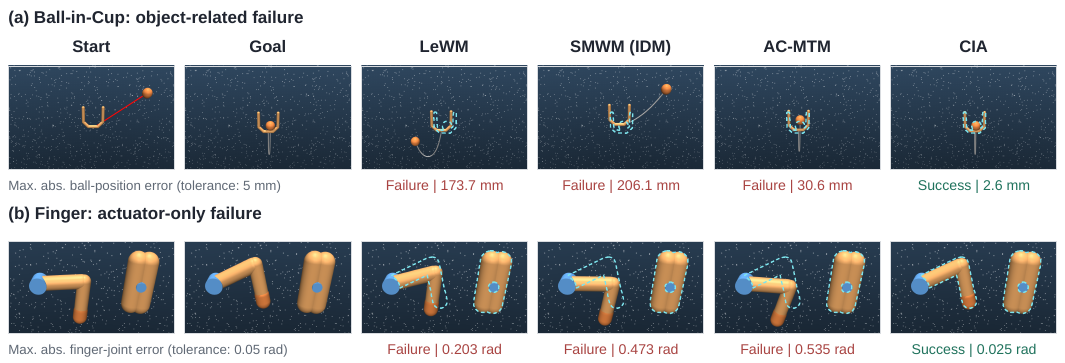}
    \caption{\textbf{Examples of planning failures.} Ball-in-Cup is shown after 25 MPPI steps, while Finger is shown after 50 CEM steps. LeWM, SMWM, and AC-MTM miss the ball-position goal (top) or finger-joint goal (bottom), whereas \method{} succeeds. Cyan dashed outlines show the goal configuration.}
    \label{fig:planning_failures}
    \vskip -0.1in
\end{figure*}

\subsection{Quality of the Learned Proposal and Score}
\label{sec:proposal_score}

The proposal supplies alternative actions, while the score learns to identify the executed action using the history and next latent.
We evaluate both at the end of training on held-out validation windows from three training seeds in the four environments with a learned proposal; Reacher uses the known uniform behavior sampler.

The learned proposal supplies alternatives adapted to the behavior data. Its samples have limited within-history dispersion and approximately reproduce aggregate action-boundary frequencies (\Cref{tab:proposal_score}). Sample variance averages within-history candidate variances in train-standardized action coordinates. Likelihood fitting adapts these candidates to the conditional behavior; the fresh probes in \Cref{sec:ablation} and Appendix~\ref{app:history_only_details} measure the identification cues that remain without the next latent.
The trained score uses the next latent to identify the executed action: accuracy is $57.6$--$89.7\%$ among $K=33$ candidates and falls to approximately $3.03\%$ chance when the next latent is replaced (\Cref{tab:proposal_score}). This confirms that the learned score responds to the observed transition, as intended by \method{}'s design.
\Cref{sec:ablation} examines how the choice of alternatives affects action identification and downstream planning performance.

\subsection{Analysis of Planning Failures}
\label{sec:failure_analysis}

To identify which goal requirements improve, we decompose unsuccessful rollouts on shared test tasks. Among episodes that never achieve full success, the final state is classified as an actuator-only failure if the object meets its goal but the actuated component does not, or an object-related failure if the object misses its goal, including joint failures. These mutually exclusive categories are reported as percentages of all episodes.

\Cref{tab:failures} shows that all three inverse objectives reduce pusher-only failures on PushT, while \method{} also has the lowest block-related failure rate. The failure categories therefore distinguish improvements in actuator positioning from improvements in meeting the object goal. \method{}'s additional reduction relative to AC-MTM is modest and uncertain under resampling. Ball-in-Cup provides a clearer separation: \method{} reduces ball-related failures by $15.9\%p$ relative to AC-MTM, while cup-only failures are rare. Most evaluated tasks start with the ball inside the cup; the result primarily concerns maintaining and transporting the ball to its goal, rather than general catching ability.

Finger shows a similar pattern: \method{} mainly reduces failures where the spinner meets its goal but the finger joints do not, while changes in spinner-related failures are smaller. Here, reaching the object's target configuration leaves another part of the goal unsatisfied: the finger must also reach the requested joint configuration. This contrasts with Ball-in-Cup, where the largest reduction concerns the ball. Thus, the gains concern different missing parts of the goal across environments, rather than consistently favoring either the object or actuator. These components are not supplied as separate training targets; the decomposition locates the observed planning gains in different goal requirements across environments. \Cref{fig:planning_failures} shows selected matched rollouts illustrating both patterns.


Frozen-encoder readouts provide an additional observation on PushT (\Cref{fig:motivation} and Appendix~\ref{app:representation_diagnostic}). Using the same linear readout protocol for each frozen encoder, we test how readily goal components can be recovered without adapting the representation. Across three seeds, \method{} retains accurate pusher position while exposing block orientation more accurately than SMWM and AC-MTM. Median pusher errors are 3--6 pixels for the inverse methods versus 20--22 for LeWM; \method{} also matches LeWM's 4--5-degree median orientation error. It therefore combines accurate actuator readout with the object-orientation readout retained by LeWM. Both components are required by the PushT goal criterion. These readouts establish accessibility of goal components, but do not directly test action information beyond the conditional action mean.

\begin{table}[t]
\centering
\caption{\textbf{Failure decomposition.} Percentages of all episodes, averaged over MPPI and CEM and five seeds. Categories use the final states of episodes that never succeed. Lower is better.}
\label{tab:failures}
\small
\setlength{\tabcolsep}{3pt}
\resizebox{\columnwidth}{!}{%
\begin{tabular}{@{}llrrrrrr@{}}
\toprule
Environment & Failure type & LeWM & TS & CGS & SMWM & AC-MTM & \method{} \\
\midrule
\multirow{2}{*}{PushT} & Pusher only & 9.0 & 13.5 & 6.4 & 1.8 & 1.8 & 2.5 \\
 & Block & 13.9 & 13.9 & 13.2 & 15.6 & 13.6 & 11.0 \\
\midrule
\multirow{2}{*}{Finger} & Joints only & 39.1 & 40.8 & 41.8 & 38.7 & 39.5 & 33.5 \\
 & Spinner & 39.3 & 36.0 & 36.5 & 37.9 & 38.3 & 36.6 \\
\midrule
\multirow{2}{*}{Ball-in-Cup} & Cup only & 0.7 & 0.5 & 1.0 & 0.8 & 0.5 & 1.0 \\
 & Ball & 72.1 & 73.0 & 72.8 & 76.9 & 75.6 & 59.7 \\
\bottomrule
\end{tabular}}
\vskip -0.1in
\end{table}



\subsection{Effect of Alternative Action Sources}
\label{sec:ablation}
Candidate construction determines which action distinctions supervise the encoder. Exact behavior matching motivates history-conditioned alternatives (\Cref{prop:cinv-matched}); here we test their practical value with proposals learned from offline data. On Ball-in-Cup, Cube, and Finger, we retrain \method{} with 32 alternatives drawn from one of three sources: the uniform distribution $q_{\mathrm{unif}}=\operatorname{Unif}(\mathcal A)$ over the action space, which ignores the behavior; the empirical marginal $q_{\mathrm{marg}}=\frac{1}{M}\sum_{i=1}^{M}\delta_{a_i}$ of the $M$ training actions, which preserves the overall action distribution; and the learned proposal $q_{\mathrm{cond}}=q_\eta(\cdot\mid h)$, which additionally accounts for the current history.

\begin{figure}[t]
    \centering
    \includegraphics[width=\linewidth]{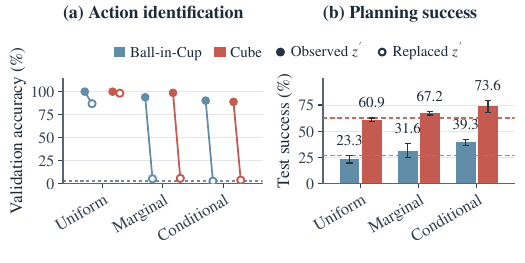}
    \vskip -0.2in
    \caption{\textbf{Alternative-action sources on Ball-in-Cup and Cube.} (a) Identification accuracy with observed (filled) or replaced (open) next latents. (b) MPPI--CEM average test success. Results show five-seed means; error bars in (b) indicate standard errors across seeds. Dashed lines mark chance in (a) and each environment's LeWM mean in (b).}
    \label{fig:negsrc_identification_planning}
    \vskip -0.2in
\end{figure}

History-conditioned alternatives achieve the highest planner-averaged mean success in all three environments. On Ball-in-Cup and Cube, they outperform marginal alternatives under bootstrap resampling (\Cref{tab:negsrc} and Appendix~\ref{app:negative_source_details}). Finger follows the same mean ordering, with a smaller, uncertain conditional--marginal gap; uniform alternatives also improve over LeWM there. These results support the practical value of history-conditioned alternatives from fitted proposals, with benefits that vary across environments and planners.


\Cref{fig:negsrc_identification_planning} separates ease of identification from planning performance. Uniform alternatives yield the highest identification accuracy but the lowest mean planning success among the three sources. Their accuracy remains high after replacing the next latent while keeping the history and candidates fixed, whereas conditional scores fall to approximately chance and marginal scores retain modestly above-chance accuracy. Conditional alternatives nevertheless yield higher mean planning success than marginal alternatives despite lower identification accuracy. These candidate sets define different classification tasks: a more easily solved task need not provide more useful representation supervision.
\paragraph{Action identification without the next latent.}
\label{sec:history_shortcut}
The replacement test examines each variant's trained score. On Ball-in-Cup and Cube, we separately fit fresh discriminators without $z'$, using candidates alone (Cand.), physical-state histories (Phys.), or frozen \method{} latent histories (Latent). These probes use the same conditional-\method{} models and data splits across sources, averaging three probe initializations and, where model-dependent, three encoder seeds. We report their held-out CE gain over chance, $\widehat{\Delta}_X=\log 33-\widehat{\mathcal L}_X$ (\Cref{tab:negsrc}). Larger gains indicate more identification cues available without the next latent; diagnostic details are in Appendix~\ref{app:history_only_details}.

Conditional sampling reduces the measured latent-history CE gain relative to marginal sampling in both environments; on Cube, it falls from $3.32$ to $1.22$ nats. Candidate-only cues remain and are stronger than with marginal alternatives (Appendix~\ref{app:history_only_details}). The fitted proposal therefore changes the available identification cues rather than eliminating them. Together with the planning ablation, these results support the practical value of learning to distinguish actions drawn for the same history, as motivated by the matched reference case in \Cref{eq:sstar}.

\begin{table}[t]
    \centering
    \caption{\textbf{Alternative-action sources.} MPPI--CEM test success is mean $\pm$ standard error over five seeds. $\widehat{\Delta}_X$ measures CE gain without the next latent.}
    \label{tab:negsrc}
    \small
    \setlength{\tabcolsep}{3pt}
    \renewcommand{\arraystretch}{1.05}
    \resizebox{\linewidth}{!}{%
    \begin{tabular}{@{}llcccc@{}}
        \toprule
        & & & \multicolumn{3}{c}{$\widehat{\Delta}_X$ (nats)}\\
        \cmidrule(lr){4-6}
        Environment & Source & Success (\%) & Cand. & Phys. & Latent\\
        \midrule
        \multirow{4}{*}{Ball-in-Cup}
        & LeWM
        & $27.1\pm0.8$ & -- & -- & --\\
        & Uniform
        & $23.3\pm1.6$ & 3.49 & 3.49 & 3.49\\
        & Marginal
        & $31.6\pm2.9$ & 0.03 & 2.27 & 1.82\\
        & Conditional $q_\eta$
        & $\mathbf{39.3}\pm1.3$ & 0.90 & 1.88 & 0.99\\
        \midrule
        \multirow{4}{*}{Cube}
        & LeWM
        & $62.7\pm0.4$ & -- & -- & --\\
        & Uniform
        & $60.9\pm0.8$ & 3.50 & 3.50 & 3.50\\
        & Marginal
        & $67.2\pm0.8$ & 0.03 & 3.33 & 3.32\\
        & Conditional $q_\eta$
        & $\mathbf{73.6}\pm2.5$ & 1.27 & 1.42 & 1.22\\
        \bottomrule
    \end{tabular}%
    }
    \vskip -0.1in
\end{table}

\section{Conclusion and Limitation}
\label{sec:limitation}



We introduced \method{}, an auxiliary objective that identifies executed actions among alternatives sampled from a history-conditioned proposal. Whereas additional information from the next latent improves optimal squared-error regression only through changes in the conditional action mean, it can lower the optimal contrastive identification loss under exact behavior matching even when that mean remains unchanged. Across five visual control benchmarks, \method{} achieves higher mean planner-averaged success than LeWM and the regression-IDM baseline without modifying the planner. These results support contrastive action identification with history-conditioned alternatives as a useful objective for offline world-model learning.

\method{} depends on the quality of its conditional action proposal. Imperfect behavior matching can leave cues that help identify the executed action without using the next latent, and likelihood fitting does not necessarily eliminate them. Moreover, \method{} learns a separate score that is not used during planning. Its action-identification objective does not directly optimize multi-step prediction accuracy.
\clearpage
\section*{LLM Usage Statement}
We used OpenAI ChatGPT, OpenAI Codex, and Anthropic Claude Code for research ideation and theoretical discussion; code development, debugging, and experiment automation; analysis and figure preparation; and language editing. The figures were generated from actual experimental data using plotting code developed with LLM assistance. The authors are responsible for the scientific claims, experimental results, citations, and all submitted content.
{
    \small
    \bibliographystyle{ieeenat_fullname}
    \bibliography{main}
}

\clearpage
\onecolumn
\appendix
\setcounter{page}{1}
\begin{center}
    {\Large\bfseries \thetitle\par}
    \vspace{0.5em}
    {\Large Supplementary Material\par}
\end{center}
\vspace{1em}

\section{Analysis of regression inverse dynamics}
\label{app:idm}

The following results concern a fixed encoder and a specified history variable $H$, assuming $\mathbb{E}\|A\|_2^2<\infty$. All infima over predictors are population quantities taken over unrestricted measurable functions with finite risk.

\subsection{Proof of Proposition~\ref{prop:idm}}

\begin{proof}
Write $m_H=\mathbb{E}[A\mid H]$ and $m_{HZ}=\mathbb{E}[A\mid H,Z']$. Conditional expectation yields the orthogonal decomposition
\begin{equation}
  \begin{aligned}
    \mathbb{E}\|A-r(H,Z')\|_2^2
    &=\mathbb{E}\|A-m_{HZ}\|_2^2\\
    &\quad+\mathbb{E}\|m_{HZ}-r(H,Z')\|_2^2,
  \end{aligned}
\end{equation}
because the cross term has zero conditional expectation given $(H,Z')$. The minimizing predictor is therefore $m_{HZ}$, giving $V_{H,Z'}=\mathbb{E}\|A-m_{HZ}\|_2^2$. By the same argument, $V_H=\mathbb{E}\|A-m_H\|_2^2$. Setting $r(H,Z')=m_H$ in the decomposition above gives
\begin{equation}
  V_H=V_{H,Z'}+\mathbb{E}\|m_{HZ}-m_H\|_2^2,
\end{equation}
which proves \eqref{eq:delta}. Since both terms on the right-hand side are nonnegative, $0\le\Delta\le V_H$. For $V>0$, defining $R_H^2=1-V_H/V$ gives $V_H=(1-R_H^2)V$. Equivalently, the total action variance decomposes as
\begin{equation}
  V=R_H^2V+\Delta+V_{H,Z'},
  \label{eq:idm-decomposition}
\end{equation}
into the history-explained component, the additional contribution of $Z'$, and the remaining prediction error.
\end{proof}

\subsection{Interpretation and Qualifications}
\label{app:idm-interpretation}

\paragraph{Matching the conditioning variables.}
\Cref{prop:idm} compares both regression problems using the same history $H$. If an inverse head receives only a restricted history summary $\chi(H)$ alongside $Z'$, its history-only baseline is determined by $\chi(H)$ rather than $H$, and its gain from observing $Z'$ can exceed $\Delta$. For example, if $A$ is determined by $H$, $\chi$ is constant, and $Z'$ reveals $A$, then $\Delta=0$ for the full history even though adding $Z'$ to $\chi(H)$ reduces the minimum error from $V$ to zero.

\paragraph{Mean-preserving action information.}
By \eqref{eq:delta}, $\Delta=0$ if and only if
\begin{equation}
\mathbb E[A\mid H,Z']=\mathbb E[A\mid H]
\quad\text{almost surely}.
\label{eq:idm-zero-gain}
\end{equation}
However, equal conditional means do not imply equal conditional distributions. Take $H$ constant, $A$ uniform on $\{-2,-1,1,2\}$, and $Z'=|A|$. Given either value of $Z'$, the two possible actions have opposite signs and equal probability, so
\begin{equation}
\mathbb E[A\mid H]=\mathbb E[A\mid H,Z']=0,
\qquad
V_H=V_{H,Z'}=\frac{5}{2}.
\label{eq:idm-magnitude-risk}
\end{equation}
Yet $Z'$ reveals the action magnitude exactly: writing $\mathsf H$ for Shannon entropy,
\begin{equation}
I(A;Z'\mid H)
=\mathsf H(Z'\mid H)-\mathsf H(Z'\mid H,A)
=\log 2.
\label{eq:idm-magnitude-information}
\end{equation}
Squared-error regression does not reward this information, whereas \Cref{cor:cinv-dependence} and the calculation in \Cref{app:cinv-magnitude-example} show that behavior-matched contrastive learning does.

\paragraph{Scope.}
This analysis applies specifically to squared-error inverse regression at a fixed encoder and history $H$; a probabilistic inverse model can also respond to distributional changes beyond the conditional mean. The distinct role of behavior matching in \method{} is to eliminate history-only discrimination under the ideal proposal. Because these results are population-level and fix the representation, they do not directly characterize finite-sample optimization, encoder gradients, or downstream planning. \Cref{app:inverse_head_diagnostics} complements this analysis with empirical measurements of history-only predictability and next-latent regression gains on frozen benchmark encoders.

\section{Analysis of conditional inverse contrastive learning}
\label{app:cinv}

Fix the encoder and the conditioning variable $H$. We assume regular conditional distributions and a common action-space reference measure (which may include boundary atoms for censored actions) under which $p_\beta(\cdot\mid h,z')$, $p_\beta(\cdot\mid h)$, and $q(\cdot\mid h)$ admit densities, with $p_\beta(\cdot\mid h)\ll q(\cdot\mid h)$ almost surely. Alternatives are drawn conditionally independently from $q(\cdot\mid H)$ and are independent of $(A_0,Z')$ given $H$. All logarithms are natural, and $K\ge2$ counts all candidates. Infima over scores are taken over unrestricted measurable candidate-wise functions, allowing limiting sequences when zero posterior probabilities require diverging logits.

\subsection{Random-index representation and optimal score}
\label{app:cinv-score}

Let $J$ be uniform on $\{0,\ldots,K-1\}$, independent of $(H,A_0,Z',A_1,\ldots,A_{K-1})$. Form the candidate tuple $C=(C_0,\ldots,C_{K-1})$ by placing the executed action at $C_J=A_0$ and the alternatives at the remaining positions. For a shared candidate-wise score $S$, define
\begin{equation}
  \pi_S(j\mid h,z',c)
  =\frac{\exp S(h,z',c_j)}{\sum_i\exp S(h,z',c_i)}.
\end{equation}
Since jointly permuting candidates and their label leaves the loss unchanged,
\begin{equation}
  \mathcal{L}_{\mathrm{CIA}}(S;q)
  =\mathbb{E}[-\log\pi_S(J\mid H,Z',C)].
  \label{eq:cinv-index-loss}
\end{equation}
Let
\begin{equation}
  \rho(h,z',a)=\frac{p_\beta(a\mid h,z')}{q(a\mid h)}.
\end{equation}
Conditioned on $(H,Z',J)=(h,z',j)$, the candidate density is $p_\beta(c_j\mid h,z')\prod_{i\ne j}q(c_i\mid h)$. By Bayes' rule,
\begin{equation}
  \pi^*(j\mid h,z',c)
  :=\mathbb{P}(J=j\mid h,z',c)
  =\frac{\rho(h,z',c_j)}{\sum_i\rho(h,z',c_i)}.
  \label{eq:cinv-index-posterior}
\end{equation}
Writing $\mathsf H$ for Shannon entropy, the cross-entropy decomposes as
\begin{equation}
  \begin{aligned}
    \mathcal{L}_{\mathrm{CIA}}(S;q)
    &=\mathsf H(J\mid H,Z',C)\\
    &\quad+\mathbb{E}\operatorname{KL}
      \bigl(\pi^*(\cdot\mid H,Z',C)
      \,\|\,\pi_S(\cdot\mid H,Z',C)\bigr).
  \end{aligned}
  \label{eq:cinv-ce-decomposition}
\end{equation}
Hence $S^*=\log\rho+c(h,z')$ is Bayes-optimal wherever finite, defined up to an action-independent additive term $c(h,z')$. Substituting $p_\beta(a\mid h,z')=p(z'\mid h,a)p_\beta(a\mid h)/p(z'\mid h)$ proves \eqref{eq:sstar}. Here $p(z'\mid h,a)$ is the conditional density under the data distribution; if $h$ does not determine the full state, it may still depend on the behavior policy through the unobserved state. When the posterior assigns zero probability to some candidates, the optimum requires $-\infty$ logits, which finite neural scores can approach in the limit. This recovers the standard density-ratio interpretation of contrastive classification \citep{oord2018representation,ma-collins-2018-noise}.

\subsection{Proof of Proposition~\ref{prop:cinv-matched}}
\label{app:cinv-matched-proof}

\begin{proof}
Using the random-index construction of \Cref{app:cinv-score} with positive position $J$ and candidate tuple $C$, exact matching ($q=p_\beta$) implies that marginalizing out $Z'$ renders all $K$ candidates i.i.d.\ given $H$. Thus $J$ is uniform given $(H,C)$, and
\begin{equation}
\mathsf H(J\mid H,C)=\log K.
\end{equation}
Any history-only score has cross-entropy at least this entropy, proving \eqref{eq:cinv-history-floor}. By \eqref{eq:cinv-ce-decomposition}, every score $S$ satisfies
\begin{equation}
\begin{aligned}
\log K-\mathcal L_{\mathrm{CIA}}(S;p_\beta)
&\le\mathsf H(J\mid H,C)-\mathsf H(J\mid H,Z',C)\\
&=I(J;Z'\mid H,C).
\end{aligned}
\end{equation}
Because $A_0=C_J$ is determined by $(J,C)$ and $(J,C)$ is conditionally independent of $Z'$ given $(H,A_0)$, we have
\begin{equation}
I(J,C;Z'\mid H)=I(A_0;Z'\mid H).
\label{eq:cinv-information-identity}
\end{equation}
Applying the chain rule gives $I(J;Z'\mid H,C)\le I(J,C;Z'\mid H)=I(A_0;Z'\mid H)=I(A;Z'\mid H)$, proving \eqref{eq:cinv-cmi}. Note that \eqref{eq:cinv-information-identity} holds for any general $q$ satisfying the same conditional-independence sampling assumptions.
\end{proof}

\subsection{Proof of Corollary~\ref{cor:cinv-dependence}}
\label{app:cinv-corollary-proof}

We first show that conditional dependence is equivalent to a strictly positive optimal contrastive gain, using the optimal-score characterization rather than the mutual-information bound alone.

\begin{proof}
Under exact behavior matching, define the optimal contrastive gain
\begin{equation}
G_K:=\log K-\inf_S\mathcal L_{\mathrm{CIA}}(S;p_\beta).
\label{eq:cinv-optimal-gain}
\end{equation}
Since the density-ratio score realizes the Bayes posterior in \eqref{eq:cinv-index-posterior} (in the limiting sense above), $\inf_S\mathcal L_{\mathrm{CIA}}(S;p_\beta)=\mathsf H(J\mid H,Z',C)$. Combined with $\mathsf H(J\mid H,C)=\log K$, this gives
\begin{equation}
G_K=I(J;Z'\mid H,C).
\label{eq:cinv-gain-identity}
\end{equation}
We claim that
\begin{equation}
G_K=0
\quad\Longleftrightarrow\quad
A\perp\!\!\!\perp Z'\mid H.
\label{eq:cinv-zero-gain}
\end{equation}
If $A\perp\!\!\!\perp Z'\mid H$, then the density ratio is identically one, the posterior over $J$ is uniform, and $G_K=0$.

Conversely, suppose $G_K=0$, so the posterior of $J$ given $(H,Z',C)$ is uniform almost surely. For almost every $(h,z')$, write $P_h=p_\beta(\cdot\mid h)$ and
\begin{equation}
\rho(a)=\frac{p_\beta(a\mid h,z')}{p_\beta(a\mid h)},
\qquad
\int\rho(a)\,P_h(\mathrm da)=1.
\end{equation}
Relative to $Q=P_h^{\otimes K}$, the candidate law given $(h,z')$ has density $\overline\rho(c)=K^{-1}\sum_i\rho(c_i)$. Where $\overline\rho(c)>0$, a uniform posterior requires all $\rho(c_i)$ to be equal; where $\overline\rho(c)=0$, nonnegativity implies they are all zero. Thus $\rho(C_0)=\rho(C_1)$ holds $Q$-almost surely. Since $C_0$ and $C_1$ are i.i.d.\ under $Q$, $\rho$ must be constant $P_h$-almost surely, and unit expectation forces $\rho=1$. Therefore $p_\beta(\cdot\mid h,z')=p_\beta(\cdot\mid h)$, establishing \eqref{eq:cinv-zero-gain}.

Under the hypotheses of \Cref{cor:cinv-dependence}, equal conditional means imply $\Delta=0$ by \Cref{prop:idm}, while $I(A;Z'\mid H)>0$ rules out conditional independence and hence gives $G_K>0$ by \eqref{eq:cinv-zero-gain}. Equivalently, $\inf_S\mathcal L_{\mathrm{CIA}}(S;p_\beta)<\log K$.
\end{proof}

\paragraph{Contrastive loss in the magnitude example.}
\label{app:cinv-magnitude-example}
Returning to the example in \Cref{app:idm-interpretation} with $q=p_\beta$, a candidate has nonzero posterior probability if and only if its magnitude equals $Z'$. All such candidates share a density ratio of $2$, and their total count (including the executed action) is
\begin{equation}
N=1+B,
\qquad B\sim\operatorname{Binomial}(K-1,1/2).
\end{equation}
The Bayes posterior is uniform over these $N$ candidates, yielding
\begin{equation}
\begin{aligned}
\inf_S\mathcal L_{\mathrm{CIA}}(S;p_\beta)
&=\mathbb E[\log N],\\
G_K&=\log K-\mathbb E[\log N]>0.
\end{aligned}
\label{eq:cinv-magnitude-gain}
\end{equation}
The inequality is strict for all $K\ge2$ because $N\le K$ and $\mathbb P(N<K)>0$. For $K=2$, $G_2=\tfrac12\log 2$, whereas the regression gain remains $\Delta=0$. Dropping magnitude from the representation leaves optimal regression error unchanged while eliminating contrastive gain. The richer direction--distance example, which compares two informative encodings rather than a constant history, is detailed in \Cref{app:direction_distance_example}.

\paragraph{Limits of the comparison.}
\Cref{cor:cinv-dependence} characterizes the population capacity of the matched objective at a fixed representation. In practice, finite neural scores may not capture every dependence, finite-$K$ gains are capped at $\log K$ (and need not rank representations by mutual information), and imperfect proposals introduce the history-only bias $b(q)$ analyzed next. Goal relevance and planning utility are evaluated empirically in our experiments.

\subsection{Exact proposal-mismatch correction}
\label{app:cinv-mismatch}

Define the behavior-prior bias
\begin{equation}
  b(q):=\log K-\inf_s\mathcal{L}_{\mathrm{CIA}}(s;q),
  \label{eq:cinv-bias}
\end{equation}
where $s(h,a)$ ranges over unrestricted scores that ignore $Z'$. This measures the largest reduction below $\log K$ achievable from history and candidates alone.

\begin{proposition}[Correction for proposal mismatch]
\label{prop:cinv-mismatch}
For every score $S$,
\begin{equation}
  I(A;Z'\mid H)
  \ge\log K-\mathcal{L}_{\mathrm{CIA}}(S;q)-b(q).
  \label{eq:cinv-corrected-bound}
\end{equation}
Moreover, $0\le b(q)\le\log K$, with $b(q)=0$ if and only if $q(\cdot\mid H)=p_\beta(\cdot\mid H)$ almost surely.
\end{proposition}

\begin{proof}
Let $\rho_0(h,a)=p_\beta(a\mid h)/q(a\mid h)$. Repeating the posterior calculation without $Z'$ yields
\begin{equation}
  \mathbb{P}(J=j\mid H=h,C=c)
  =\frac{\rho_0(h,c_j)}{\sum_i\rho_0(h,c_i)}.
\end{equation}
Since $\log\rho_0$ realizes this posterior,
\begin{equation}
  \begin{aligned}
    \inf_s\mathcal{L}_{\mathrm{CIA}}(s;q)
    &=\mathsf H(J\mid H,C),\\
    b(q)&=I(J;C\mid H).
  \end{aligned}
  \label{eq:cinv-bias-information}
\end{equation}
The bounds $0\le b(q)\le\log K$ follow immediately from the entropy of $J$. Under exact matching, the posterior is uniform and $b(q)=0$. Conversely, if $b(q)=0$, the posterior is uniform and all candidates have equal $\rho_0$ almost surely. Conditioning on $H=h,J=0$, the independent variables $C_0\sim p_\beta(\cdot\mid h)$ and $C_1\sim q(\cdot\mid h)$ share the same density ratio almost surely, forcing $\rho_0(h,\cdot)$ to be constant; normalization $\int\rho_0(h,a)\,q(\mathrm da\mid h)=1$ sets this constant to $1$, so $q=p_\beta$ almost surely.

Finally, applying \eqref{eq:cinv-information-identity} (which holds for general $q$) yields
\begin{equation}
  \begin{aligned}
    &\log K-b(q)-\mathcal{L}_{\mathrm{CIA}}(S;q)\\
    &\quad\le\mathsf H(J\mid H,C)-\mathsf H(J\mid H,Z',C)\\
    &\quad=I(J;Z'\mid H,C)\le I(A;Z'\mid H).
  \end{aligned}
\end{equation}
\end{proof}

\paragraph{KL upper bound on the bias.}
Define
\begin{equation}
  D_\beta(q):=\mathbb{E}_H\operatorname{KL}
  \bigl(p_\beta(\cdot\mid H)\,\|\,q(\cdot\mid H)\bigr).
\end{equation}
We show that $b(q)\le D_\beta(q)$. Fix $h$, let $P_{j,h}$ be the candidate law given $H=h,J=j$, let $M_h=K^{-1}\sum_jP_{j,h}$, and let $Q_h=q(\cdot\mid h)^{\otimes K}$. When the divergence is finite,
\begin{equation}
  \begin{aligned}
    \operatorname{KL}(p_\beta(\cdot\mid h)\|q(\cdot\mid h))
    &=\frac1K\sum_j\operatorname{KL}(P_{j,h}\|Q_h)\\
    &=I(J;C\mid H=h)+\operatorname{KL}(M_h\|Q_h).
  \end{aligned}
\end{equation}
Taking expectations over $H$ establishes $b(q)\le D_\beta(q)$ (trivially so when $D_\beta(q)=\infty$), which gives
\begin{equation}
  I(A;Z'\mid H)
  \ge\log K-\mathcal{L}_{\mathrm{CIA}}(S;q)-D_\beta(q).
  \label{eq:cinv-kl-bound}
\end{equation}
For fixed history features and finite expectations, $\mathcal{L}_q(\eta)=D_\beta(q_\eta)+\mathrm{const}$. Maximum-likelihood fitting of $q_\eta$ therefore directly minimizes an upper bound on the history-only discrimination bias within the proposal family.

\paragraph{Relation to existing contrastive bounds.}
The density-ratio view and unconditional InfoNCE bound are well known \citep{oord2018representation,ma-collins-2018-noise,poole2019variational}. The conditional construction and KL correction in \eqref{eq:cinv-kl-bound} follow \citet{pmlr-v139-sordoni21a}, mapping their $(x,y,x')$ to $(Z',A,H)$. We do not claim this conditional InfoNCE bound as a new estimator; rather, we apply it to compare inverse objectives for offline world-model learning, where $b(q)$ gives a tighter operational correction than $D_\beta(q)$.

\subsection{Interpretation and connection to the experiments}
\label{app:cinv-scope}

\paragraph{Conditioning and scope.}
The guarantees above apply to the specified history $H$ at a fixed encoder, rather than the full raw observation history. Both the proposal and score use this same conditioning, and alternatives are sampled independently of the executed action and next observation given $H$ (see \Cref{app:cinv-implementation}).

\paragraph{Interpreting fitted probes.}
A proposal's bias depends on how well it matches the data-collection behavior: uniform sampling matches uniformly random behavior, and marginal sampling matches history-independent behavior. For any fitted history-only score $\widehat s$,
\begin{equation}
  \log K-\mathcal{L}_{\mathrm{CIA}}(\widehat s;q)\le b(q).
\end{equation}
Thus, a positive probe gain confirms outcome-blind discrimination, whereas a small gain does not prove its absence. Because a fitted probe only lower-bounds $b(q)$ (up to finite-sample variance), subtracting it from the trained score's gain does not certify a valid lower bound on $I(A;Z'\mid H)$.

\paragraph{Matched and fitted proposals.}
Under exact behavior matching, no history-only score can beat chance, forcing the inverse task to rely on $Z'$. For a general proposal, \Cref{eq:sstar} decomposes the optimal score into transition compatibility and the behavior-prior ratio, while $b(q)$ quantifies the maximal history-only gain. Fitting $q_\eta$ by maximum likelihood minimizes the forward-KL upper bound $D_\beta(q_\eta)$ within the proposal family, though it does not guarantee $b(q_\eta)=0$. Our learned-proposal experiments evaluate this practical construction against the theoretical reference of exact matching (which Reacher realizes directly via its known uniform sampler).

\paragraph{Relation to probabilistic inverse dynamics.}
The conditional-mean limitation applies to squared-error regression (as in SMWM), not to probabilistic inverse models, which can also capture distributional structure beyond the mean (\Cref{tab:toy_fixed_heads}). However, \method{} differs from likelihood-based inverse models in how it supervises the representation. A likelihood-based inverse model rewards all action predictability—including predictability from history alone, which accounts for most action variance in three of our five environments (\Cref{app:inverse_head_diagnostics}). In contrast, \method{} offloads history-based behavior modeling to a separate proposal fitted on detached features and trains the shared encoder via relative identification, where beating chance under exact matching requires information from $Z'$. Furthermore, \method{}'s cross-entropy is bounded, whereas continuous-action log-likelihood can diverge as distributions concentrate.
\Cref{app:smwm_history} examines this distinction by evaluating a probabilistic inverse-dynamics baseline trained jointly with the world model, using the same history and target transition as \method{}. While this baseline improves average planning success over regression-based inverse dynamics, \method{} retains higher planner-averaged success across all five environments.

\paragraph{Connection to planning.}
Our analyses and experiments address complementary pieces of the picture: \Cref{cor:cinv-dependence} and the fixed-encoder check confirm that action identification rewards information beyond conditional means; the candidate-source ablation tests the practical benefit of history-conditioned alternatives; and state readouts and failure decomposition examine goal-component accessibility. Together, these findings establish \method{} as effective auxiliary supervision for planning, though benchmark gains likely reflect both representation cues and changes in candidate coverage and optimization dynamics.

\section{Direction--Distance Example}
\label{app:direction_distance_example}

This section details the setup, analytical optima, and fixed-encoder learning check for \Cref{fig:cinv_information_reward}, testing whether trained neural heads reproduce the extra reward \method{} assigns to distance information.

\subsection{Construction and analytical comparison}
\label{app:cinv-control-example}

Fix the initial state and history. The action is $A=(D,B)$ with independent components $D\sim\operatorname{Unif}\{-1,1\}$ and $B\sim\operatorname{Unif}\{-2,-1,1,2\}$, producing endpoint $X'=D|B|$. We compare two fixed zero-mean, unit-variance scalar encodings:
\begin{equation}
    Z'_{\mathrm{dir}}=\operatorname{sgn}(X')=D,
    \qquad Z'_{\mathrm{full}}=X'/\sqrt{5/2}.
    \label{eq:toy-fixed-encodings}
\end{equation}
$Z'_{\mathrm{dir}}$ merges near and far endpoints in each direction, whereas $Z'_{\mathrm{full}}$ distinguishes all four endpoints. Under either encoding, the sign of $B$ remains equally likely, so
\begin{equation}
    \mathbb E[A\mid Z'_{\mathrm{dir}}]
    =\mathbb E[A\mid Z'_{\mathrm{full}}]=(D,0),
    \qquad
    V_{H,Z'_{\mathrm{dir}}}=V_{H,Z'_{\mathrm{full}}}=5/2.
\end{equation}
Here MSE sums both raw action-coordinate errors. Both encodings reveal direction and reduce regression error from $7/2$ to $5/2$, isolating the incremental value of distance once direction is known: $\mathbb{E}[A\mid D,|B|]=\mathbb{E}[A\mid D]=(D,0)$, yet $I(A;|B|\mid D)=\log 2$. Unlike the zero-gain scalar example for \Cref{cor:cinv-dependence}, the total regression gain over the fixed history is nonzero. Alternatives are sampled from the exact uniform behavior distribution over all eight actions.

For \method{}, the candidate set comprises the executed action and $K-1$ independent draws with replacement from the uniform behavior distribution (retaining duplicates), jointly permuted with the positive label. Compatible alternatives match direction under $Z'_{\mathrm{dir}}$ ($p=1/2$) and both direction and magnitude under $Z'_{\mathrm{full}}$ ($p=1/4$). The Bayes posterior is uniform over the $1+\operatorname{Binomial}(K-1,p)$ compatible candidates, giving optimal loss
\begin{equation}
    \ell_K(p)
    =\sum_{j=0}^{K-1}\binom{K-1}{j}p^j(1-p)^{K-1-j}\log(1+j).
    \label{eq:toy-compatible-loss}
\end{equation}
Coupling the Bernoulli draws via shared uniforms shows $\ell_K(1/4)<\ell_K(1/2)$ for every $K\ge2$. At $K=33$, the optimal losses are $2.818752$ and $2.156962$ nats (with zero posteriors handled via the limiting-score convention in \Cref{app:cinv-score}). A categorical probabilistic inverse model likewise benefits, reducing its optimal NLL from $\log4$ to $\log2$.

To connect these encodings to planning, suppose a goal endpoint is sampled uniformly from $\{-2,-1,1,2\}$ and provided only through its encoding. Under $Z'_{\mathrm{dir}}$, two equally likely distances map to the same representation, capping any controller's average success at $1/2$. Under $Z'_{\mathrm{full}}$, exact latent dynamics and exhaustive action search achieve $100\%$ success.

\subsection{Fixed-encoder neural learning check}
\label{app:cinv_toy_learning}

We train only the inverse heads on these fixed encodings. Separate regression, contrastive, and eight-class probabilistic heads each use a two-hidden-layer GELU MLP of width 64. The regression and categorical heads take the latent $Z'$ as input, while the contrastive score takes $(Z', a)$. We train with AdamW (learning rate $10^{-3}$, weight decay $10^{-4}$, cosine decay) for 2,500 updates across five paired initializations and evaluate final checkpoints without checkpoint selection. Regression MSE and categorical NLL are averaged exactly over all eight transitions per update, whereas the contrastive head uses 256 balanced candidate sets per update with $K=33$.

At evaluation, MSE and categorical NLL are computed exactly over all eight transitions, while contrastive CE is estimated on 262,144 independently sampled, transition-balanced candidate sets shared across encodings and initializations. \Cref{tab:toy_fixed_heads} compares learned losses with population optima, and \Cref{fig:cinv_information_reward} plots the regression and \method{} results.

\begin{table}[t]
    \centering
    \caption{\textbf{Fixed-encoder learning check.} Learned values are means across five initializations; MSE uses raw coordinates, and CE/NLL are in nats.}
    \label{tab:toy_fixed_heads}
    \small
    \setlength{\tabcolsep}{7pt}
    \begin{tabular}{@{}llcc@{}}
        \toprule
        Encoding & Metric & Learned & Theory\\
        \midrule
        Direction only & IDM MSE & 2.500000 & 2.500000\\
        & \method{} CE & 2.819141 & 2.818752\\
        & Probabilistic IDM NLL & 1.386317 & 1.386294\\
        \midrule
        Direction + distance & IDM MSE & 2.500000 & 2.500000\\
        & \method{} CE & 2.158911 & 2.156962\\
        & Probabilistic IDM NLL & 0.693300 & 0.693147\\
        \bottomrule
    \end{tabular}
\end{table}

Learned \method{} CE closely matches theory, with sample standard deviations of $0.000005$ and $0.000276$ (smaller than the marker size in \Cref{fig:cinv_information_reward}). On the same evaluation bank, the oracle CE values are $2.819075$ and $2.158121$, so the trained scores exceed the finite-bank optima by only $0.000066$ and $0.000791$ nats. Overall, neural heads faithfully reproduce the theoretical objective gap: regression is blind to distance, whereas both \method{} and probabilistic IDM strongly reward it.

\subsection{Separate joint-learning check}
\label{app:cinv_toy_joint}

The fixed-encoder check holds the representation fixed. This check also learns the encoder, so prediction and the inverse objective shape the representation together.

\paragraph{Setup.}
Observations are $224\times224$ RGB images that show the object and an independent visual distractor. We use 512 training, 128 validation, and 512 test visual contexts. Each context is paired with all eight actions, and the distractor stays fixed within each transition. The encoder applies two stride-two $3\times3$ convolutions with 8 and 16 channels, followed by fully connected layers of widths 64 and 8, with GELU activations between layers. The forward predictor and the inverse heads are MLPs with two hidden layers of 64 GELU units.

\paragraph{Objectives and training.}
We compare four variants. All share prediction MSE and SIGReg~\citep{balestriero2025lejepa} with weight $0.09$, 128 random unit projections, and 17 knots on $[0,3]$. Three variants add one inverse loss with coefficient $0.1$: regression IDM, \method{}, or probabilistic IDM (categorical NLL). The encoder receives gradients from every active loss. The predictor receives only prediction gradients, and the prediction target is not detached. We train for 3,000 updates with batch size 128, AdamW (learning rate $3\times10^{-4}$, weight decay $10^{-4}$, cosine decay). Initialization, data order, and regularizer randomness are paired across variants within each of five initializations.

\paragraph{Evaluation and results.}
A ridge probe on the frozen encoder reads out distance, with regularization selected on validation contexts. One-step planning enumerates all eight actions with the learned predictor and picks the action with the smallest latent goal distance; it succeeds only if it reaches the exact requested endpoint. On test contexts, all four variants reach $100\%$ planning success in all five initializations. Mean distance-readout $R^2$ is $0.9878$ for prediction with SIGReg, $0.9985$ for regression IDM, $0.9882$ for \method{}, and $0.9912$ for probabilistic IDM. This check therefore shows no \method{} advantage: in this simple setting, the shared prediction objective already preserves distance. The objective-level distinction rests on the fixed-encoder check above, and neither check shows that information beyond conditional means causes the benchmark planning gains.

\section{Environments and Datasets}
\label{app:env_dataset}

\begin{figure}[t]
    \centering
    \includegraphics[width=\linewidth]{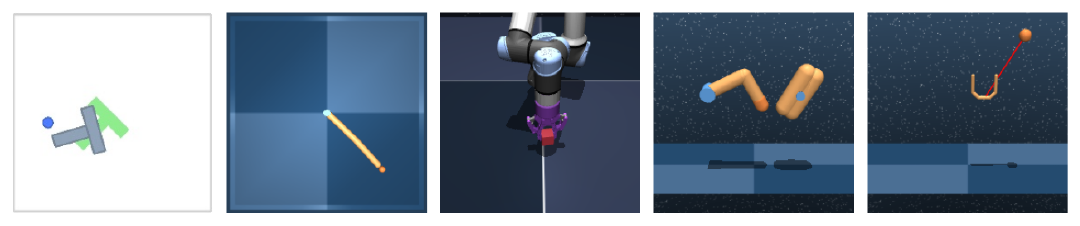}
    \caption{\textbf{Evaluation environments.} From left to right: PushT, Reacher, OGBench-Cube, Finger (turn\_easy), and Ball-in-Cup (catch).}
    \label{fig:appendix_domains}
\end{figure}

All five environments use $224\times224$ RGB observations and continuous actions (\Cref{fig:appendix_domains}). World models are trained solely on offline image--action trajectories without rewards or task labels. Environment interfaces follow PushT~\citep{chi2025diffusion}, OGBench~\citep{park2025ogbench}, and the DeepMind Control Suite~\citep{tassa2018deepmind}. We use the PushT expert, OGBench-Cube single-cube expert, and Reacher random-action datasets released with LeWM~\citep{maes2026leworldmodel}, and collect our own offline datasets for Finger (noisy heuristic policy) and Ball-in-Cup (trained soft actor-critic (SAC) policy~\citep{pmlr-v80-haarnoja18b}). Physical states are used strictly to initialize and score evaluation episodes and to run the state diagnostics.

\paragraph{Data splits and preprocessing.}
\label{app:training-preprocessing}
We use fixed, episode-disjoint train/validation/test splits shared across all methods: $16{,}817/934/934$ episodes for PushT, $9{,}000/500/500$ for Reacher and Cube, and $7{,}200/400/400$ for Finger and Ball-in-Cup. Action normalization statistics are computed on the training split and reused at evaluation. Recorded PushT actions are clipped coordinate-wise to $[-1,1]$ prior to standardization; all other datasets already lie within $[-1,1]$. During planning, MPPI and CEM optimize unconstrained sequences in standardized coordinates and unstandardize them for execution. Reacher, Finger, and Ball-in-Cup clip executed actions to $[-1,1]$, whereas PushT and Cube pass raw unstandardized actions directly to their environment interfaces.

\paragraph{Analysis of the collected datasets.}
Each of our Finger and Ball-in-Cup datasets comprises 8,000 episodes of 200 primitive steps (1.6M transitions and 1.608M RGB frames at $224\times224$ resolution), summarized in \Cref{tab:collected_dataset_statistics}. Here, reward-positive episodes contain at least one step with positive environment reward—a collection-time diagnostic distinct from our visual goal-reaching evaluation.

\begin{table}[t]
    \centering
    \small
    \caption{\textbf{Statistics of our collected datasets.}
    Action standard deviations are computed in raw coordinates over
    finite primitive-action rows, excluding terminal padding.}
    \label{tab:collected_dataset_statistics}
    \begin{tabular}{lcc}
        \toprule
        Statistic & Finger & Ball-in-Cup \\
        \midrule
        Episodes & 8,000 & 8,000 \\
        Primitive transitions & 1,600,000 & 1,600,000 \\
        Recorded frames & 1,608,000 & 1,608,000 \\
        Action std. (two coordinates)
        & $(0.359,0.683)$ & $(0.288,0.421)$ \\
        Reward-positive episodes (\%) & 53.3 & 100.0 \\
        Reward-positive final frames (\%) & 49.5 & 100.0 \\
        \bottomrule
    \end{tabular}
\end{table}

In Finger, we add independent Gaussian noise ($\sigma=0.2$) to each coordinate of the heuristic policy and clip executed actions to $[-1,1]$. The median cumulative absolute spinner rotation (summed over unwrapped angle increments) is $148.1^\circ$ per episode, with 10th and 90th percentiles of $27.3^\circ$ and $245.5^\circ$.

In Ball-in-Cup, we sample actions stochastically from an SAC policy trained on the environment's sparse catch reward ($+1$ when the ball is inside the cup target region, $0$ otherwise). Every episode achieves the catch reward at least once, with a median time to first catch of 15 primitive steps; $95.1\%$ of non-initial steps receive the catch reward, and only $5.5\%$ of episodes exhibit a positive-to-nonpositive reward transition. Post-catch stabilization therefore accounts for most of the dataset.

\paragraph{Visual goals and success criteria.}
Each visual goal-reaching task provides an initial frame and a goal frame taken 24 primitive steps later from the same held-out trajectory (where one primitive step includes any internal environment action repetition). We evaluate on 200 validation tasks (four 50-task sets) and 150 test tasks (three 50-task sets), shared across all methods and seeds. On Finger and Ball-in-Cup, start--goal pairs that already satisfy the success criterion at step 0 are filtered out prior to sampling.

An episode succeeds if its environment-specific state criterion is met at any step within the 50-primitive-step budget: concatenated pusher-and-block Euclidean position error below $20\,\mathrm{px}$ and wrapped block-orientation error below $20^\circ$ on PushT; cube-position error below $40\,\mathrm{mm}$ on Cube; every joint-coordinate error below $0.05\,\mathrm{rad}$ on Reacher; every finger-joint and spinner-hinge error below $0.05\,\mathrm{rad}$ on Finger; and every cup and ball position coordinate error below $5\,\mathrm{mm}$ on Ball-in-Cup. DMC comparisons use recorded generalized coordinates without angle wrapping. Note that failure decomposition classifies only unsuccessful episodes using their final-state errors, rather than the any-time success rule.

\paragraph{Code and data availability.}
We will release all training and evaluation code, experiment configurations, task sets, episode splits, and the Finger and Ball-in-Cup datasets along with their collection scripts.

\section{Implementation and Hyperparameters}
\label{app:hyperparams}

\subsection{Shared training configuration}
\Cref{tab:hp_shared} and \Cref{tab:hp_env} list the shared training configuration (following the LeWM backbone and schedule~\citep{maes2026leworldmodel}) and environment-specific auxiliary weights. At the model's temporal resolution, each action $a_t$ concatenates five consecutive primitive actions, giving $d_a=5d_{\mathrm{prim}}$ (10 on PushT, Reacher, Finger, and Ball-in-Cup; 25 on Cube) for both the forward predictor and all inverse objectives.

For standalone \method{}, we select $\lambda$ by validation planning performance over $\{0.01,0.03\}$ on PushT and $\{0.01,0.03,0.1,0.3,1,2,3,5\}$ on Cube. On Reacher, preliminary sweeps with a learned conditional proposal over $\{0.01,0.03,0.1,0.3,1\}$ selected $\lambda=1$; we keep $\lambda=1$ when switching to the fixed-uniform proposal on Reacher and carry it directly to Finger and Ball-in-Cup without further tuning. For \method{}+CGS on Cube, a single-seed sweep over $\lambda_{\mathrm{CIA}}\in\{0.01,0.03,0.1\}$ with $\lambda_{\mathrm{CGS}}=0.5$ selected $\lambda_{\mathrm{CIA}}=0.01$. For baselines, we adopt published auxiliary settings: LeWM-based TS and CGS from \citet{fu2026cgs}, environment-specific SMWM weights from \citet{ivashkov2026sensorimotor}, and the fixed AC-MTM weight and temperature from \citet{boylan2026no}; on Finger and Ball-in-Cup, each baseline reuses its Reacher weights. Because auxiliary objectives differ in scale, numerical weights are not directly comparable across methods.

\begin{table}[htbp]
  \centering
  \small
  \caption{\textbf{Shared training configuration.} Backbone, prediction, and regularization settings are common across methods.}
  \label{tab:hp_shared}
  \begin{tabular}{@{}ll@{}}
    \toprule
    Encoder            & ViT-Tiny, patch size 14, $224\times224$ input \\
    Latent             & 192-d (projector MLP, hidden width 2048) \\
    Predictor          & causal transformer, depth 6, 16 heads, MLP width 2048, dropout 0.1 \\
    History            & 3 frames \\
    Temporal sampling  & 5 primitive actions per model step \\
    SIGReg             & weight $\alpha=0.09$, 1024 projections, 17 knots \\
    Optimizer          & AdamW~\citep{loshchilov2019decoupled}, learning rate $5\times10^{-5}$, weight decay $10^{-3}$ \\
    Learning-rate schedule & linear warm-up over the first 1\% of updates, cosine decay to 0 \\
    Gradient clipping  & norm 1.0 \\
    Batch size         & 128 \\
    Precision          & bfloat16 mixed \\
    Updates            & 127{,}960 (139{,}290 on PushT) \\
    Seeds              & 5 per environment \\
    \bottomrule
  \end{tabular}
\end{table}

\begin{table}[htbp]
  \centering
  \small
  \caption{\textbf{Environment-specific auxiliary weights.} The first row gives the primitive action dimension.}
  \label{tab:hp_env}
  \begin{tabular}{@{}lccccc@{}}
    \toprule
                     & Reacher & PushT & Cube & Ball-in-Cup & Finger \\
    \midrule
    Primitive dimension $d_{\mathrm{prim}}$ & 2   & 2    & 5   & 2   & 2   \\
    TS weight        & 0.1 & 0.01 & 0.1 & 0.1 & 0.1 \\
    CGS weight       & 0.5 & 0.01 & 0.5 & 0.5 & 0.5 \\
    SMWM (IDM) weight & 5  & 30   & 1   & 5   & 5   \\
    AC-MTM (action-NCE) weight & 0.30 & 0.30 & 0.30 & 0.30 & 0.30 \\
    \method{} weight $\lambda$ & 1 & 0.01 & 3 & 1  & 1   \\
    \method{}+CGS: $\lambda_{\mathrm{CIA}}$ & 1 & 0.01 & 0.01 & 1 & 1 \\
    \method{}+CGS: $\lambda_{\mathrm{CGS}}$ & 0.5 & 0.01 & 0.5 & 0.5 & 0.5 \\
    \bottomrule
  \end{tabular}
\end{table}

\subsection{Baseline auxiliary objectives}
TS and CGS operate on all four latents in each training window, with auxiliary weights linearly ramped over the first 5\% of updates~\citep{fu2026cgs}. CGS uses standardized actions; on Cube, it uses the 15 translation coordinates of five-primitive-step blocks where the gripper is commanded closed. Following \citet{ivashkov2026sensorimotor}, SMWM predicts each action from its adjacent latent pair $(z_t,z_{t+1})$ and averages inverse MSE across the three pairs in each window. To keep non-collapse regularization consistent across methods, we include SIGReg (weight $0.09$) in both SMWM and AC-MTM. \Cref{app:smwm_history} additionally evaluates SMWM-H, which supplies the regression head with the exact same causal history and query as \method{}.

For AC-MTM~\citep{boylan2026no}, an inverse MLP maps each adjacent latent pair to an action query $u_i$ and computes logits $-\|u_i-a_j\|_2^2/(0.10\,d_a)$ against all normalized actions in the batch (384 candidates per query for batch size 128). The total loss is $\mathcal L_{\mathrm{pred}}+0.30\,\mathcal L_{\mathrm{action\text{-}NCE}}+0.09\,\mathcal L_{\mathrm{SIGReg}}$, updating the encoder and inverse head without a direct gradient path to the forward predictor. Because AC-MTM and \method{} differ in score parameterization, history inputs, and candidate count as well as candidate distribution, \Cref{app:negative_source_details} isolates candidate-source effects within \method{}. (On Reacher, uniform history-independent behavior makes marginal and conditional action distributions identical.)

\subsection{\method{} implementation}
\label{app:cinv-implementation}

Both the score and the learned proposal condition on the same causal history,
\begin{equation}
    h_t = \bigl(z_{t-2}, a_{t-2}, z_{t-1}, a_{t-1}, z_t\bigr),
\end{equation}
comprising three observation latents and two intervening actions, with next latent $z'=z_{t+1}$. Neither $a_t$ nor $z'$ enters $h_t$. To prevent target-frame leakage through BatchNorm statistics in the auxiliary branch, the shared projector uses running statistics snapshotted before the current update while preserving gradients through the encoder and projector. Architectural details appear in \Cref{tab:hp_cinv}.

\begin{table}[htbp]
  \centering
  \small
  \caption{\textbf{\method{} components.} The proposal and score are used only during training.}
  \label{tab:hp_cinv}
  \begin{tabular}{@{}ll@{}}
    \toprule
    Candidates $K$ & 33 (the executed action and 32 alternatives) \\
    Behavior model $q_\eta(a\mid h)$ & MLP, 2 hidden layers of 256 units, GELU \\
                        & 5-component diagonal Gaussian mixture over the action, \\
                        & censored at the action bounds \\
    Proposal optimizer & AdamW, learning rate $10^{-3}$, no weight decay, gradient clipping 1.0 \\
    Score $S_\psi(h,z',a)$   & MLP, 2 hidden layers of 256 units, GELU, temperature 1 \\
    Score optimizer    & shared world-model optimizer and schedule \\
    \bottomrule
  \end{tabular}
\end{table}

\paragraph{Proposal fitting and update order.}
Starting from random initialization on PushT, Cube, Finger, and Ball-in-Cup, we first prefit $q_\eta$ for up to 5{,}000 updates with the world model frozen, selecting the checkpoint by validation NLL. The proposal is a 5-component diagonal Gaussian mixture censored at the action bounds: samples outside the bounds are clipped, so its log-likelihood combines interior densities with boundary point masses. At each joint training step, we sample and detach alternatives from the pre-update proposal, update $q_\eta$ by NLL on detached history features, and then perform the world-model update using the stored candidates. Reacher uses a fixed-uniform proposal matching its known behavior and skips proposal fitting.

\paragraph{Contrastive update.}
All $K=33$ candidates are scored against the same observed $z'$, without requiring counterfactual rollouts. Candidate positions and the positive index are jointly permuted, and $S_\psi$ directly outputs logits with no $-\log q_\eta$ correction or external temperature. The contrastive loss has constant weight $\lambda$ and updates only the score and the shared encoder/projector (via both $h_t$ and $z'$), leaving the forward predictor and proposal untouched by \method{} gradients. Latent prediction and SIGReg proceed unchanged. Only the encoder and forward predictor are retained for planning.

\subsection{Planning configuration}
All methods share the same terminal latent goal cost and planner settings within each environment (\Cref{tab:hp_planner}). Although the forward predictor is trained on three-frame histories, each planning call initializes its rollout from the current observation alone with an empty past-action context, appending candidate actions and predicted latents autoregressively (up to the three most recent pairs). This procedure runs at steps 0 and 25.

\begin{table}[htbp]
  \centering
  \small
  \caption{MPPI and CEM settings shared across methods within each environment. The maximum execution budget is 50 primitive environment steps; plans are computed at steps 0 and 25, each spanning five model steps (25 primitive steps).}
  \label{tab:hp_planner}
  \begin{tabular}{@{}ll@{}}
    \toprule
    MPPI & 128 samples, 30 iterations, temperature 64 (4 on PushT), initial standard deviation 1.0 \\
    CEM  & 128 samples, 30 iterations, 32 elites, initial standard deviation 1.0 \\
    \bottomrule
  \end{tabular}
\end{table}

\subsection{Sensitivity to the Number of Action Candidates}
\label{app:k_sweep}

We use $K=33$ (1 executed action and 32 alternatives) as our default across all environments. To test sensitivity to candidate count, we sweep $K\in\{9,17,33,65,129\}$ while holding all other training and planning settings fixed, evaluating five seeds per setting on the same 150 test tasks for both MPPI and CEM. As \Cref{tab:k_sweep} shows, reducing $K$ below 33 hurts performance—most notably on Finger—whereas increasing $K$ to 65 or 129 yields no consistent gain. Thus, $K=33$ strikes a strong balance between planning accuracy and scoring cost.

\begin{table}[h]
    \centering
    \small
    \setlength{\tabcolsep}{6pt}
    \caption{\textbf{Sensitivity to the number of action candidates $K$.} MPPI--CEM average success (\%), reported as mean $\pm$ SE across five training seeds. The $K=33$ column reuses the main-experiment results.}
    \label{tab:k_sweep}
    \begin{tabular}{lccccc}
        \toprule
        & \multicolumn{5}{c}{Total number of candidates $K$} \\
        \cmidrule(lr){2-6}
        Environment
        & 9 & 17 & \textbf{33} & 65 & 129 \\
        \midrule
        PushT
        & $85.8 \pm 0.8$
        & $85.7 \pm 0.8$
        & $86.5 \pm 0.6$
        & $87.0 \pm 0.7$
        & $86.6 \pm 0.7$ \\
        Reacher
        & $75.2 \pm 0.6$
        & $74.7 \pm 0.3$
        & $74.7 \pm 0.7$
        & $75.5 \pm 0.7$
        & $75.3 \pm 0.3$ \\
        OGBench-Cube
        & $68.8 \pm 2.5$
        & $73.1 \pm 0.3$
        & $73.6 \pm 2.5$
        & $70.3 \pm 2.9$
        & $69.4 \pm 3.3$ \\
        Finger
        & $24.9 \pm 1.2$
        & $28.3 \pm 0.5$
        & $29.9 \pm 0.8$
        & $29.6 \pm 0.9$
        & $29.7 \pm 1.6$ \\
        Ball-in-Cup
        & $36.0 \pm 2.2$
        & $33.7 \pm 2.6$
        & $39.3 \pm 1.3$
        & $36.0 \pm 3.2$
        & $39.9 \pm 0.3$ \\
        \bottomrule
    \end{tabular}
\end{table}

\subsection{Diagnostic probe configuration}
\label{app:probe-implementation}
Outcome-blind probes in \Cref{app:history_only_details} use a three-hidden-layer MLP of width 512 trained with AdamW (learning rate $10^{-3}$, no weight decay, batch size 512, cosine decay) for up to 20,000 updates, with validation-CE early stopping (patience 5,000 updates). All reported comparisons share this exact budget.

\section{Candidate Identification without the Next Latent}
\label{app:history_only_details}

\paragraph{Protocol.}
To quantify how easily alternatives can be distinguished without observing the outcome $Z'$, we freeze three conditional-\method{} encoders per environment on Ball-in-Cup and Cube and train fresh outcome-blind discriminators on 33 candidates drawn from each source in \Cref{sec:ablation} (using samples from the frozen model's learned $q_\eta$ for the conditional source). Note that these diagnostics share the same frozen conditional-\method{} encoders across sources, whereas the planning ablations in \Cref{app:negative_source_details} retrain separate world models per source.

Each discriminator receives candidate actions paired with one of three contexts: no context ($X=\varnothing$), physical-state history with past actions ($X=H_{\mathrm{phys}}$), or the model's causal latent history with past actions ($X=H$). None receives $Z'$. We use 307,200 training, 10,240 validation, and 20,480 test episode-disjoint windows, sampling fresh training alternatives with randomized candidate positions and selecting checkpoints by validation CE (\Cref{app:probe-implementation}). We average three probe initializations within each encoder seed and then across the three encoder seeds (model-independent uniform and marginal candidate-only probes are averaged over three initializations once per environment).

\paragraph{Metrics and results.}
We measure held-out CE gain over chance ($\log 33 \approx 3.497$ nats; chance accuracy $1/33 \approx 3.03\%$),
\begin{equation}
    \widehat{\Delta}_X=\log33-\widehat{\mathcal L}_X.
    \label{eq:history_shortcut_gain}
\end{equation}
\Cref{tab:history_only_full} shows a clear progression across candidate sources. Uniform alternatives are nearly perfectly identifiable without any context ($\ge 99.8\%$ accuracy). Marginal sampling removes this candidate-only shortcut ($\sim 3.3\%$ accuracy), but supplying history restores high identifiability ($60.0\%$ on Ball-in-Cup and $92.9\%$ on Cube with latent history). History-conditioned sampling ($q_\eta$) substantially lowers latent-history CE gain relative to marginal sampling—dropping from $1.82$ to $1.00$ nats on Ball-in-Cup and from $3.32$ to $1.22$ nats on Cube.

On Cube, the latent-history probe scores slightly lower than the candidate-only probe for conditional alternatives ($35.8\%$ vs.\ $36.7\%$). This reflects finite-capacity optimization rather than a loss of information from adding history; a longer training run on one seed narrows but does not eliminate the gap.

\begin{table}[t]
    \centering
    \caption{\textbf{Identification without the next latent.} Held-out accuracy and CE of fresh discriminators. CE and gain are in nats; aggregation is specified above.}
    \label{tab:history_only_full}
    \small
    \setlength{\tabcolsep}{6pt}
    \begin{tabular}{@{}lllrrr@{}}
        \toprule
        Environment & Alternatives & Context & Acc. (\%) & CE & $\widehat{\Delta}_X$\\
        \midrule
        Ball-in-Cup & Uniform & Candidates only & 99.8 & 0.006 & 3.491 \\
         &  & Physical history & 99.9 & 0.003 & 3.494 \\
         &  & Latent history & 99.9 & 0.005 & 3.492 \\
         & Marginal & Candidates only & 3.3 & 3.463 & 0.033 \\
         &  & Physical history & 66.9 & 1.230 & 2.267 \\
         &  & Latent history & 60.0 & 1.676 & 1.821 \\
         & Conditional $q_\eta$ & Candidates only & 29.6 & 2.594 & 0.903 \\
         &  & Physical history & 59.3 & 1.613 & 1.884 \\
         &  & Latent history & 41.5 & 2.502 & 0.995 \\
        \midrule
        Cube & Uniform & Candidates only & 100.0 & 0.000 & 3.497 \\
         &  & Physical history & 100.0 & 0.001 & 3.495 \\
         &  & Latent history & 100.0 & 0.001 & 3.496 \\
         & Marginal & Candidates only & 3.4 & 3.471 & 0.026 \\
         &  & Physical history & 93.7 & 0.163 & 3.333 \\
         &  & Latent history & 92.9 & 0.177 & 3.320 \\
         & Conditional $q_\eta$ & Candidates only & 36.7 & 2.228 & 1.269 \\
         &  & Physical history & 41.2 & 2.076 & 1.421 \\
         &  & Latent history & 35.8 & 2.275 & 1.222 \\
        \bottomrule
    \end{tabular}
\end{table}

\paragraph{Residual proposal cues.}
While conditional sampling suppresses history-based shortcuts, it leaves measurable candidate-only cues: $29.6\%$ accuracy on Ball-in-Cup and $36.7\%$ on Cube (raising Cube's candidate-only CE gain from $0.03$ to $1.27$ nats). Thus, the learned $q_\eta$ improves upon marginal sampling in practice (\Cref{app:negative_source_details}) without achieving exact behavior matching. In \Cref{tab:proposal_score}, proposal dispersion and boundary frequencies capture specific distributional statistics rather than the full conditional law (see \Cref{app:inverse_head_diagnostics} for direct history-only $R^2$ estimates), and residual mismatches still enter \Cref{eq:sstar} via the behavior-prior ratio. Across all entries in \Cref{tab:history_only_full}, the maximum standard error of $\widehat{\Delta}_X$ is $0.031$ nats. As noted in \Cref{app:cinv-scope}, each probe's gain estimates its own performance rather than upper-bounding $b(q)$, so subtracting it from the trained score's gain does not yield a certified mutual-information bound.

\paragraph{Relation to the trained score.}
On these same three-seed test cohorts, the trained conditional-\method{} score achieves $89.3\%$ accuracy ($0.285$ nats CE) on Ball-in-Cup and $88.8\%$ ($0.339$ nats) on Cube. Replacing $Z'$ with another test window's next latent collapses accuracy to $2.8\%$ and $3.7\%$, respectively. Hence, the trained score relies strongly on $Z'$ even though fresh probes can still extract residual cues without it. Note that the slight differences among reported Ball-in-Cup accuracies ($89.7\%$ in \Cref{tab:proposal_score}, $90.3\%$ in \Cref{tab:negsrc_score_replacement}, and $89.3\%$ here) reflect their distinct evaluation cohorts: three models on validation, five models on validation, and three models on test, respectively.

\section{Representation Accessibility}
\label{app:representation_diagnostic}

We evaluate frozen PushT encoders from LeWM, SMWM, AC-MTM, and \method{} (three training seeds each) to assess how accessible physical state variables are in the learned representations.

\paragraph{Linear state readout.}
Following frozen-encoder probing in LeWM and SMWM~\citep{maes2026leworldmodel,ivashkov2026sensorimotor}, we fit ridge regression on 100,000 training frames, tune regularization on 10,000 validation frames, and evaluate on 20,000 test frames across episode-disjoint splits (standardizing features and targets using training statistics). Block angle is encoded as $(\cos\theta,\sin\theta)$ with aggregated $R^2$, and position scores average coordinate-wise $R^2$.

\begin{figure}[t]
    \centering
    \includegraphics[width=\linewidth]{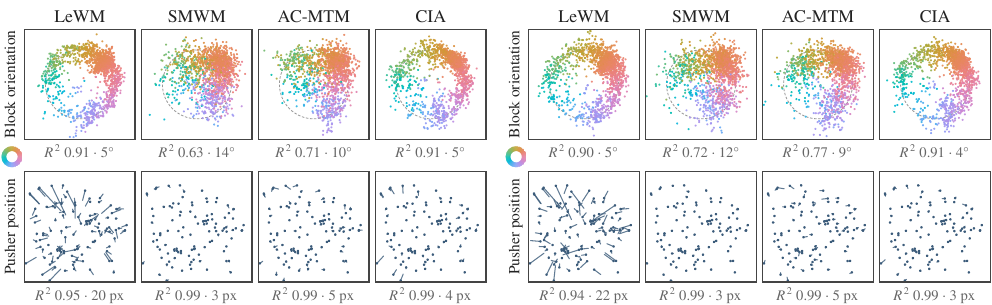}
    \caption{\textbf{Readout planes for the other two encoder seeds.} Same visualization as \Cref{fig:motivation}.}
    \label{fig:readout_planes_seeds}
\end{figure}

\Cref{tab:representation_readout} summarizes test $R^2$ across seeds. \Cref{fig:motivation} visualizes one seed per method on a shared random test subset (top: predicted block $(\cos\theta,\sin\theta)$; bottom: true vs.\ predicted pusher positions), and \Cref{fig:readout_planes_seeds} shows the remaining two seeds. Results are highly consistent across all three seeds: median pusher error is 20--22 pixels for LeWM versus 2.85--5.48 pixels for the inverse methods (rounded to the nearest integer in figure labels), while median block-angle error is 4--5$^\circ$ for LeWM and \method{}, compared to 12--14$^\circ$ for SMWM and 7--10$^\circ$ for AC-MTM.

\begin{table}[t]
    \centering
    \caption{
    \textbf{Linear state readout on PushT.} Test $R^2$, reported as mean $\pm$ sample standard deviation across three frozen encoder seeds. Higher is better.
    }
    \label{tab:representation_readout}
    \small
    \begin{tabular}{@{}lccc@{}}
        \toprule
        Method & Pusher position & Block position & Block angle \\
        \midrule
        LeWM
        & $0.9428\pm0.0061$
        & $0.974\pm0.004$
        & $0.910\pm0.007$ \\
        SMWM
        & $0.9941\pm0.0003$
        & $0.949\pm0.007$
        & $0.656\pm0.060$ \\
        AC-MTM
        & $0.9916\pm0.0002$
        & $0.959\pm0.005$
        & $0.766\pm0.055$ \\
        \method{}
        & $0.9905\pm0.0008$
        & $0.975\pm0.003$
        & $0.918\pm0.010$ \\
        \bottomrule
    \end{tabular}
\end{table}

\begin{figure}[t]
    \centering
    \includegraphics[width=\linewidth]
        {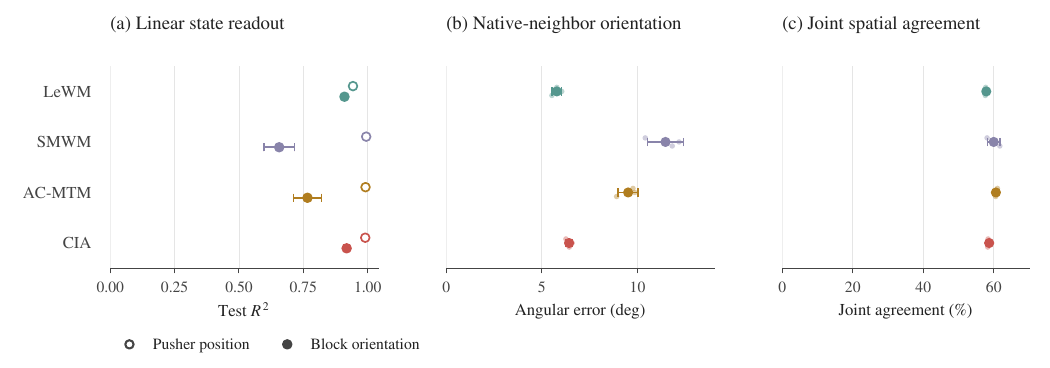}
    \caption{\textbf{Representation diagnostics on PushT.} (a) Linear readout of pusher position and block orientation. (b) Block-angle error of the retrieved frame. Each model retrieves the pool frame whose latent is nearest to the query latent in squared Euclidean distance. (c) Fraction of queries whose retrieved frame is within $20$ pixels in pusher and block position and within $20^\circ$ in block angle. Error bars show sample standard deviations across three frozen encoder seeds. Joint agreement is a retrieval diagnostic, not planning success.}
    \label{fig:representation_diagnostics}
\end{figure}

\paragraph{Interpretation.}
\method{} uniquely combines sharp pusher-position readout ($R^2=0.9905$) with strong block-orientation readout ($R^2=0.918$), matching LeWM on orientation while substantially outperforming SMWM ($0.656$) and AC-MTM ($0.766$). \Cref{fig:representation_diagnostics} pairs these linear readouts with native-latent retrieval diagnostics—measuring nearest-neighbor orientation error (panel b) and joint position-orientation agreement (panel c) under unscaled squared Euclidean latent distance. Together, these diagnostics show that \method{} preserves multiple goal-relevant state components simultaneously, even where native retrieval geometry alone is not uniformly best.

In the PushT failure decomposition (\Cref{sec:failure_analysis}), this orientation readout advantage accompanies a $2.6\%p$ higher mean block-related failure rate for \method{} relative to AC-MTM. Here, block-related failures denote unsuccessful episodes finishing with block-only Euclidean position error $\ge 20\,\mathrm{px}$ or wrapped block-orientation error $\ge 20^\circ$, expressed as a fraction of all evaluation episodes.

\section{Inverse Dynamics with Matched History}
\label{app:smwm_history}

We compare regression and probabilistic inverse supervision using the same causal history and target transition as \method{}. This examines whether matching the inverse input or replacing conditional-mean regression with action-distribution modeling recovers \method{}'s planning performance. All methods retain latent prediction and SIGReg, and planning uses the encoder and forward predictor without the auxiliary inverse head. The hyperparameter search and selection procedure for the additional probabilistic inverse-dynamics baseline is described here alongside its formulation and planning results.

\paragraph{History-matched regression.}
SMWM-H replaces SMWM's pairwise inverse input $(z_t,z')$ with $(h_t,z')$, where $h_t$ contains the same three observation latents and two intervening actions used by \method{}. The inverse head remains a two-hidden-layer ReLU MLP of width 256, with its first layer widened to accommodate the history. We retain the original SMWM loss weights: 30 on PushT, 1 on Cube, and 5 on Reacher, Finger, and Ball-in-Cup. The backbone, prediction and SIGReg losses, optimizer, training budget, data splits, and preprocessing follow the corresponding SMWM configuration in \Cref{tab:hp_shared,tab:hp_env}. Matching \method{}'s history also aligns the prediction target: SMWM-H predicts the executed action for the final transition in each window---the only one with a complete history---whereas standard SMWM averages the regression loss over all three transitions in the window. The forward predictor's history is unchanged.

\paragraph{Probabilistic inverse dynamics.}
The p-IDM baseline models $p_\phi(a_t\mid h_t,z')$ using the same history and final-transition target as SMWM-H and \method{}. Its inverse head is a two-hidden-layer GELU MLP of width 256 that parameterizes a five-component diagonal Gaussian mixture. The action distribution is censored to the valid action bounds, so the likelihood accounts for interior densities as well as boundary probability masses. Component scales have a minimum of $10^{-3}$ in standardized action coordinates. The training objective is
\begin{equation}
    \mathcal L_{\mathrm{p\text{-}IDM}}
    =
    \mathcal L_{\mathrm{pred}}
    + 0.09\,\mathcal L_{\mathrm{SIGReg}}
    + \lambda_{\mathrm{p\text{-}IDM}}
    \mathbb E\left[
        -\frac{1}{d_a}
        \log p_\phi(A\mid H,Z')
    \right],
\end{equation}
where $(A, H, Z')$ denotes the random transition tuple and $d_a$ is the model-step action dimension. Dividing by $d_a$ expresses the inverse loss in nats per action coordinate. The likelihood term updates the inverse head and shared encoder through both the history latents and the next latent, without a direct gradient path to the forward predictor. This baseline is trained jointly with the world model, unlike the frozen-encoder likelihood probes in \Cref{app:inverse_head_diagnostics}.

For the reported p-IDM configurations, we select $\lambda_{\mathrm{p\text{-}IDM}}$ from $\{0.1,1,5\}$ using the mean validation success across MPPI and CEM. The selected weights are 0.1 on PushT, and 5 on Reacher, Cube, Finger, and Ball-in-Cup. For each seed, all methods are evaluated with both planners on the same 150 test tasks, using the corresponding five-seed sets from the main experiments. We first average MPPI and CEM within each seed and then aggregate across seeds.

\begin{table}[t]
    \centering
    \caption{\textbf{Inverse dynamics with matched history.} Test success (\%), reported as mean $\pm$ SE across five training seeds. Average combines MPPI and CEM within each seed. The highest and second-highest distinct means in each row are shown in bold and underlined, respectively; ties share the same formatting.}
    \label{tab:smwm_history}
    \small
    \setlength{\tabcolsep}{5pt}
    \renewcommand{\arraystretch}{1.05}
    \begin{tabular}{@{}llcccc@{}}
        \toprule
        Environment & Planner & SMWM & SMWM-H & p-IDM & \method{} \\
        \midrule
        \multirow{3}{*}{PushT}
        & MPPI    & $76.9\pm1.2$ & $76.9\pm0.7$ & $\underline{79.9}\pm2.4$ & $\mathbf{81.9}\pm1.1$ \\
        & CEM     & $88.3\pm1.5$ & $\underline{89.3}\pm1.1$ & $\mathbf{91.1}\pm1.0$ & $\mathbf{91.1}\pm0.5$ \\
        & Average & $82.6\pm1.2$ & $83.1\pm0.8$ & $\underline{85.5}\pm1.0$ & $\mathbf{86.5}\pm0.6$ \\
        \midrule
        \multirow{3}{*}{Reacher}
        & MPPI    & $65.2\pm0.6$ & $65.1\pm0.5$ & $\mathbf{67.5}\pm0.4$ & $\underline{67.3}\pm1.0$ \\
        & CEM     & $80.4\pm1.2$ & $79.3\pm1.9$ & $\underline{80.5}\pm1.3$ & $\mathbf{82.0}\pm0.8$ \\
        & Average & $72.8\pm0.8$ & $72.2\pm0.8$ & $\underline{74.0}\pm0.8$ & $\mathbf{74.7}\pm0.7$ \\
        \midrule
        \multirow{3}{*}{Cube}
        & MPPI    & $68.1\pm0.7$ & $68.0\pm0.9$ & $\underline{72.9}\pm0.7$ & $\mathbf{74.7}\pm2.3$ \\
        & CEM     & $69.2\pm0.4$ & $67.5\pm0.9$ & $\underline{70.4}\pm1.0$ & $\mathbf{72.5}\pm2.7$ \\
        & Average & $68.7\pm0.3$ & $67.7\pm0.8$ & $\underline{71.7}\pm0.7$ & $\mathbf{73.6}\pm2.5$ \\
        \midrule
        \multirow{3}{*}{Finger}
        & MPPI    & $19.1\pm0.9$ & $\underline{19.7}\pm0.5$ & $\mathbf{20.9}\pm0.5$ & $16.7\pm0.9$ \\
        & CEM     & $27.7\pm2.0$ & $30.0\pm0.9$ & $\underline{31.6}\pm0.7$ & $\mathbf{43.2}\pm1.1$ \\
        & Average & $23.4\pm1.1$ & $24.9\pm0.6$ & $\underline{26.3}\pm0.4$ & $\mathbf{29.9}\pm0.8$ \\
        \midrule
        \multirow{3}{*}{Ball-in-Cup}
        & MPPI    & $\underline{15.3}\pm0.8$ & $14.1\pm1.5$ & $13.5\pm1.6$ & $\mathbf{20.7}\pm1.6$ \\
        & CEM     & $29.2\pm1.6$ & $\underline{36.4}\pm1.4$ & $34.5\pm2.5$ & $\mathbf{58.0}\pm2.0$ \\
        & Average & $22.3\pm0.9$ & $\underline{25.3}\pm0.8$ & $24.0\pm1.9$ & $\mathbf{39.3}\pm1.3$ \\
        \bottomrule
    \end{tabular}
\end{table}

\paragraph{Planning results.}
Matching the regression input and prediction target does not close the mean planning gap to \method{} (\Cref{tab:smwm_history}). The equally weighted overall means are $53.9\%$ for SMWM, $54.6\%$ for SMWM-H, and $60.8\%$ for \method{}. For the \method{}--SMWM-H comparison, a bootstrap over shared tasks and independently resampled training seeds within each method, using 5,000 bootstrap replicates, yields a 95\% percentile interval of $[4.0,8.4]$ percentage points for the overall gain of $6.2$ percentage points.
Probabilistic inverse supervision improves planner-averaged mean success over SMWM in all five environments and over SMWM-H in four, reaching an overall mean of $56.3\%$. It approaches \method{} on PushT and Reacher, matches its CEM mean on PushT, and achieves higher MPPI means on Reacher and Finger. Nevertheless, \method{} retains the highest planner-averaged mean in every environment, with the largest differences on Ball-in-Cup and Finger. On Ball-in-Cup---where the gap is largest---using $\lambda_{\mathrm{p\text{-}IDM}}=0.1$ instead of the validation-selected weight of 5 yields $24.9\pm0.8\%$ planner-averaged success (compared with $24.0\pm1.9\%$ in \Cref{tab:smwm_history}), still well below \method{}.

\paragraph{Relation to the objective.}
The p-IDM comparison extends the evaluation beyond conditional-mean regression: a probabilistic inverse objective can capture higher-order structure and multimodality in the conditional action distribution, yet this capability alone does not reproduce \method{}'s mean planning performance. The objectives differ in how they supervise the shared representation. p-IDM directly maximizes the likelihood of the executed action given the history and next latent, whereas \method{} separates behavior modeling from distinguishing the executed action among history-conditioned alternatives. Under exact behavior matching, performing this identification above chance requires information in the next latent beyond what is predictable from the history. The planning results support the practical value of this design alongside forward prediction, while showing that probabilistic inverse supervision is itself a competitive alternative in several environments.

\begin{figure}[t]
    \centering
    \includegraphics[width=\linewidth]{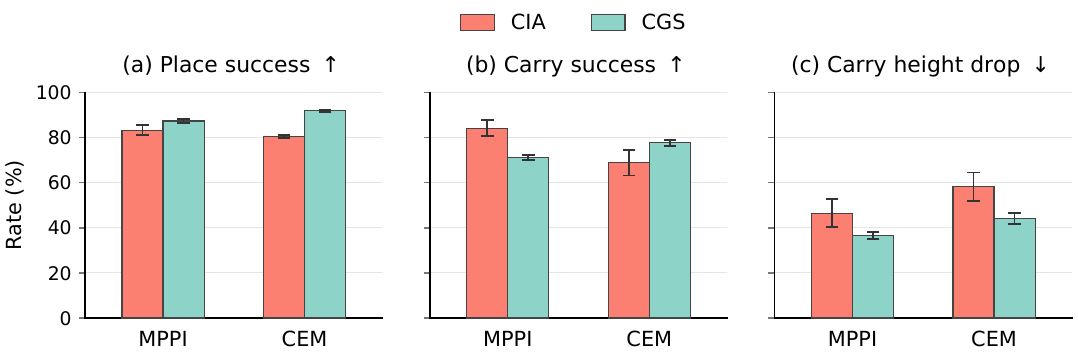}
    \caption{\textbf{Carrying and placement on Cube.} Mean $\pm$ SE across five training seeds, with 93 place and 34 carry tasks per seed. Success is measured over 50 steps; height drop denotes the cube crossing below 3\,cm during the first 25 steps of carry tasks.}
    \label{fig:cube_carry_place}
\end{figure}

\begin{figure}[t]
    \centering
    \includegraphics[width=\linewidth]{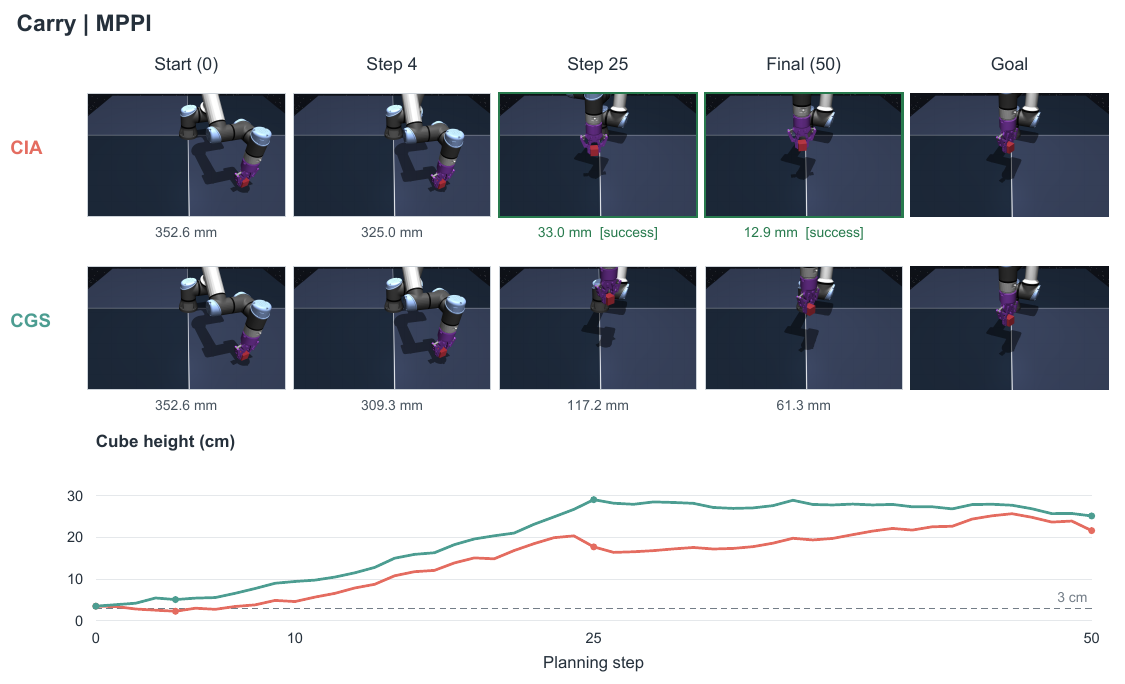}
    \caption{\textbf{Carry: height drop need not imply failure.} Matched MPPI rollouts with the same start and goal. Under CIA, the cube briefly drops below the 3\,cm height threshold but then reaches the goal; under CGS, the cube stays elevated but misses the target. Numbers show cube-position error; green borders indicate error within the 40\,mm success threshold. Curves show cube height over time.}
    \label{fig:cube_carry_example}
\end{figure}

\begin{figure}[t]
    \centering
    \includegraphics[width=\linewidth]{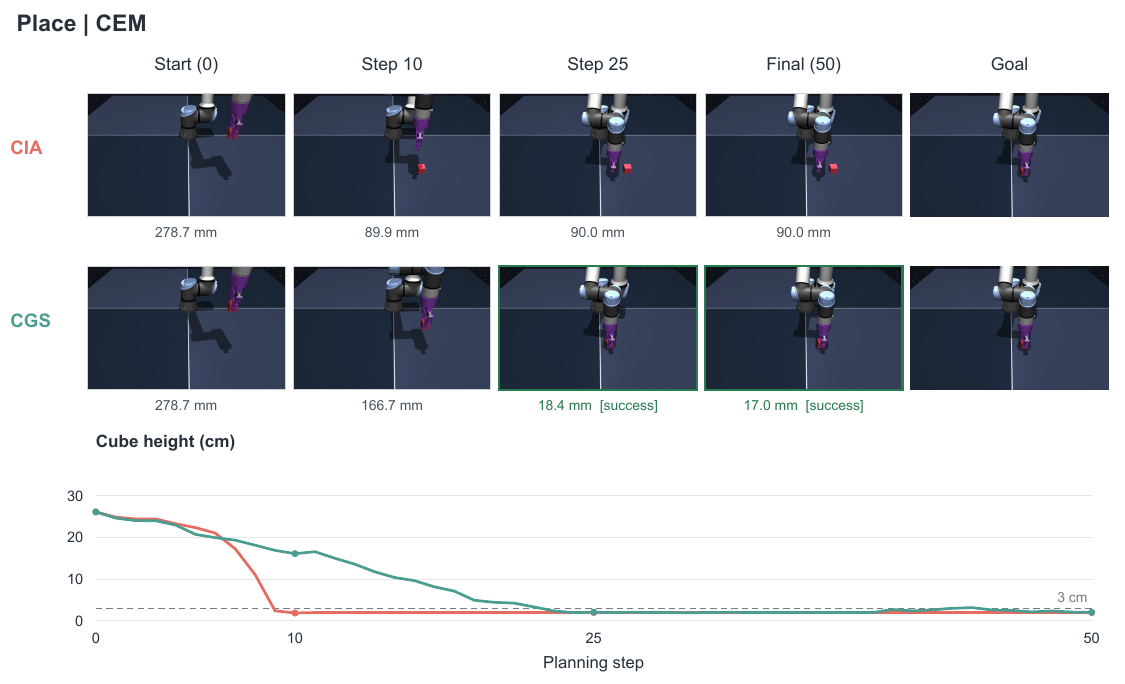}
    \caption{\textbf{Place: lowering the cube is not sufficient.} Matched CEM rollouts with the same start and goal. Both methods lower the cube, but only CGS reaches the target position. Numbers show cube-position error; green borders indicate error within the 40\,mm success threshold. Curves show cube height over time, with the dashed line marking 3\,cm.}
    \label{fig:cube_place_example}
\end{figure}

\section{Diagnostics of Planning Geometry on Cube}
\label{app:cube_planning_geometry}

To understand how learned latent goal costs shape action search on Cube, we compare frozen \method{} and CGS models from \Cref{tab:planning_mppi_cem} across planning budgets, gradient sensitivities, and execution patterns.

First, we vary one planning-budget parameter at a time on 200 validation tasks per seed. Expanding MPPI's sample count from 32 to 512 (at 30 iterations) narrows the CGS--\method{} mean success gap from $5.4\%p$ to $1.9\%p$, with most of the gain realized by 128 samples. Under gradient-based planning (GD), however, \method{} rises from $66.1\%$ (10 iterations) to $72.2\%$ (30 iterations) and then plateaus at $72.7\%$ (200 iterations), whereas CGS climbs steadily from $66.3\%$ to $79.1\%$. Thus, \method{} benefits from broader sampling but gains little from extended local gradient refinement.

\paragraph{Action sensitivity of the planning cost.}
To inspect which action coordinates receive optimization signal, we record latent goal-cost gradients over 100 iterations of the first 25-step plan (one seed, 350 held-out tasks: 200 validation and 150 test). Summing squared normalized-action gradients over iterations and time steps, we compute each coordinate's share per task and average across tasks. In \method{}, gripper and yaw coordinates account for only $1.15\%$ and $0.04\%$ of the total squared gradient, compared with $7.16\%$ and $2.31\%$ in CGS. \method{}'s planning gradients are therefore heavily concentrated on translation, leaving weaker relative signals for grasp and orientation adjustments. Notably, CGS's higher gripper and yaw sensitivity emerges indirectly, as its geometric alignment term supervises only translation coordinates.

\paragraph{Failures during carrying and placement.}
We analyze MPPI and CEM rollouts across 350 held-out tasks per seed, defining an elevated cube as height $>3\,\mathrm{cm}$ and requiring start--goal displacement $>40\,\mathrm{mm}$. This yields 34 \emph{carry} tasks (elevated start and goal) and 93 \emph{place} tasks (elevated start, goal below $3\,\mathrm{cm}$). As \Cref{fig:cube_carry_place} shows, CGS achieves higher mean place success with both planners, whereas carry performance splits by planner: \method{} leads under MPPI, and CGS leads under CEM. During the first 25 steps of carry tasks, \method{} allows the cube to dip below $3\,\mathrm{cm}$ more frequently, though temporary height drops do not always cause failure (\Cref{fig:cube_carry_example}); conversely, lowering the cube in place tasks does not guarantee reaching the target (\Cref{fig:cube_place_example}).

\paragraph{Relation to the objective.}
\method{} supervises the encoder so that actions are identifiable by the inverse score, but planning relies on the forward predictor and Euclidean latent goal distance. Information preserved for action identification need not map evenly onto latent distances or propagate cleanly through multi-step rollouts, because \method{} neither updates the predictor directly nor constrains the geometry of the planning cost. The narrowing MPPI gap, GD plateau, and translation-dominated gradients all point to this distinction—motivating the combination of action identification (\method{}) with geometric regularization (CGS).

\begin{table}[t]
    \centering
    \caption{\textbf{Planner-specific alternative-source ablation.} Test success (\%), mean $\pm$ standard error across five seeds. Average is computed within each seed.}
    \label{tab:negsrc_planners}
    \small
    \setlength{\tabcolsep}{4pt}
    \begin{tabular}{@{}lccc@{}}
        \toprule
        Method / alternatives & MPPI & CEM & Average \\
        \midrule
        \multicolumn{4}{@{}l}{Ball-in-Cup ($\lambda=1$)} \\
        LeWM
        & $16.1\pm0.6$ & $38.1\pm1.6$ & $27.1\pm0.8$ \\
        SMWM
        & $15.3\pm0.8$ & $29.2\pm1.6$ & $22.3\pm0.9$ \\
        \method{}, uniform
        & $17.7\pm1.3$ & $28.9\pm2.9$ & $23.3\pm1.6$ \\
        \method{}, marginal
        & $17.9\pm2.0$ & $45.3\pm5.4$ & $31.6\pm2.9$ \\
        \method{}, conditional
        & $\mathbf{20.7}\pm1.6$ & $\mathbf{58.0}\pm2.0$ & $\mathbf{39.3}\pm1.3$ \\
        \midrule
        \multicolumn{4}{@{}l}{Cube ($\lambda=3$)} \\
        LeWM
        & $61.7\pm0.5$ & $63.6\pm0.4$ & $62.7\pm0.4$ \\
        SMWM
        & $68.1\pm0.7$ & $69.2\pm0.4$ & $68.7\pm0.3$ \\
        \method{}, uniform
        & $60.0\pm1.1$ & $61.9\pm0.7$ & $60.9\pm0.8$ \\
        \method{}, marginal
        & $68.1\pm0.9$ & $66.3\pm0.9$ & $67.2\pm0.8$ \\
        \method{}, conditional
        & $\mathbf{74.7}\pm2.3$ & $\mathbf{72.5}\pm2.7$ & $\mathbf{73.6}\pm2.5$ \\
        \midrule
        \multicolumn{4}{@{}l}{Finger ($\lambda=1$)}\\
        LeWM & $14.8\pm1.0$ & $28.4\pm0.7$ & $21.6\pm0.7$\\
        SMWM & $19.1\pm0.9$ & $27.7\pm2.0$ & $23.4\pm1.1$\\
        \method{}, uniform & $\mathbf{20.0}\pm0.5$ & $32.7\pm1.4$ & $26.3\pm0.5$\\
        \method{}, marginal & $18.9\pm0.8$ & $36.7\pm1.1$ & $27.8\pm0.7$\\
        \method{}, conditional & $16.7\pm0.9$ & $\mathbf{43.2}\pm1.1$ & $\mathbf{29.9}\pm0.8$\\
        \bottomrule
    \end{tabular}
\end{table}

\section{Additional Results for Alternative Action Sources}
\label{app:negative_source_details}

Here we report planner-specific results for the Ball-in-Cup, Cube, and Finger candidate-source ablations (\Cref{sec:ablation}). Each \method{} variant uses five training seeds and 32 alternatives with all other hyperparameters, initializations, and data orders held fixed, evaluated on the same 150 test tasks (25-step horizon, replanning once for a 50-step budget). LeWM and SMWM are included as reference baselines. Under the task-and-seed bootstrap of \Cref{app:smwm_history}, the conditional-minus-marginal gain has 95\% intervals of $[1.2,14.5]$ on Ball-in-Cup, $[0.1,12.1]$ on Cube, and $[-1.3,5.7]$ on Finger (\%p).

\begin{table}[t]
    \centering
    \caption{\textbf{Next-latent replacement.} Validation identification accuracy (\%), averaged over five trained models per variant. Chance is $3.03\%$.}
    \label{tab:negsrc_score_replacement}
    \small
    \setlength{\tabcolsep}{6pt}
    \begin{tabular}{@{}lcc@{}}
        \toprule
        Alternatives & Observed $z'$ & Replaced $z'$ \\
        \midrule
        \multicolumn{3}{@{}l}{Ball-in-Cup} \\
        Uniform & 100.0 & 86.9 \\
        Marginal & 93.9 & 5.4 \\
        Conditional $q_\eta$ & 90.3 & 2.8 \\
        \midrule
        \multicolumn{3}{@{}l}{Cube} \\
        Uniform & 100.0 & 98.3 \\
        Marginal & 98.7 & 5.9 \\
        Conditional $q_\eta$ & 88.9 & 3.9 \\
        \bottomrule
    \end{tabular}
\end{table}

History-conditioned alternatives achieve the highest planner-averaged mean success in all three environments (\Cref{tab:negsrc_planners}). On Ball-in-Cup and Cube, conditional sampling outperforms both uniform and marginal alternatives across both planners, beating marginal sampling on all five Ball-in-Cup seeds and four of five Cube seeds. On Finger, conditional sampling likewise beats marginal sampling on all five seeds in the planner average, driven by a strong CEM gain ($43.2\%$ vs.\ $36.7\%$) despite uniform sampling leading under MPPI ($20.0\%$ vs.\ $16.7\%$).

\paragraph{Dependence of the trained score on the next latent.}
To test whether each variant's trained score actually relies on $z'$, we evaluate validation accuracy before and after randomly permuting $z'$ across valid queries in the same batch (excluding self-matches, with histories and candidates fixed). As \Cref{tab:negsrc_score_replacement} shows, uniform-alternative scores remain highly accurate even with mismatched $z'$ ($86.9\%$ on Ball-in-Cup, $98.3\%$ on Cube), confirming severe outcome-blind shortcuts. In contrast, conditional scores drop to chance ($2.8\%$ and $3.9\%$), and marginal scores retain only modest above-chance accuracy ($5.4\%$ and $5.9\%$). Notably, conditional sampling yields the lowest observed-$z'$ classification accuracy yet the highest planning success—demonstrating that discriminating harder, history-plausible alternatives provides more useful representation supervision than achieving high accuracy on easy negatives.

\section{Action Predictability from History and the Next Latent}
\label{app:inverse_head_diagnostics}

To link \Cref{prop:idm} to our benchmark encoders, we measure how strongly actions are predictable from history alone, how much $Z'$ adds under regression and probabilistic objectives, and whether these predictive gains track planning success. We freeze the main-experiment encoders (LeWM, SMWM, AC-MTM, and \method{}) alongside the jointly trained p-IDM encoders from \Cref{app:smwm_history}, and train separate regression and probabilistic inverse heads taking either $H$ or $(H,Z')$.

\paragraph{Protocol.}
All methods share episode-disjoint sets of 204,800 training windows (sampled from 40\% of training episodes), 10,240 validation windows, and 20,480 test windows per environment, using \method{}'s three-frame causal history $H$. We evaluate three encoder seeds with two probe initializations each. Every head is a three-hidden-layer GELU MLP (width 1024, dropout 0.1) trained for 20,000 updates (batch size 1024, AdamW at learning rate $10^{-3}$ with cosine decay), with standardized inputs and validation-loss checkpointing. Probabilistic heads use five-component diagonal Gaussian mixtures censored at the action bounds. We average the two probe initializations per encoder and report mean $\pm$ SE across the three encoder seeds.

\paragraph{Metrics.}
On the test set, let $\widehat V$ be the total standardized action variance, and let $\widehat M_H$ and $\widehat M_{HZ}$ be the coordinate-summed regression errors. We report
\begin{equation}
    \widehat R_H^2 = 1-\frac{\widehat M_H}{\widehat V},
    \qquad
    \frac{\widehat\Delta}{\widehat V}
    =\frac{\widehat M_H-\widehat M_{HZ}}{\widehat V},
    \qquad
    \widehat G_{\mathrm{NLL}}
    =\widehat{\mathcal L}_{\mathrm{NLL},H}
    -\widehat{\mathcal L}_{\mathrm{NLL},HZ}.
\end{equation}
Here, $\widehat\Delta/\widehat V$ is the empirical counterpart of the mean-shift term in \Cref{prop:idm} (sensitive only to conditional means), whereas $\widehat G_{\mathrm{NLL}}$ (in nats per action vector: $d_a=25$ on Cube, $10$ elsewhere) is the empirical counterpart of $I(A;Z'\mid H)$ and also captures changes in spread, modality, and boundary mass. Because these two metrics live on different scales, we compare encoders within each metric rather than across them.

\begin{table}[t]
    \centering
    \caption{\textbf{Action predictability on frozen representations.} Test results are mean $\pm$ SE across three encoder seeds, after averaging two probe initializations within each encoder. SEs displayed as $0.000$ are below $0.0005$.}
    \label{tab:inverse_head_diagnostics}
    \small
    \setlength{\tabcolsep}{6pt}
    \renewcommand{\arraystretch}{1.05}
    \begin{tabular}{@{}llccc@{}}
        \toprule
        Environment & Encoder & $\widehat R_H^2$ & $\widehat\Delta/\widehat V$ & $\widehat G_{\mathrm{NLL}}$ (nats) \\
        \midrule
        \multirow{5}{*}{PushT}
        & LeWM      & $0.840\pm0.000$ & $0.070\pm0.000$ & $1.657\pm0.035$ \\
        & SMWM      & $0.830\pm0.001$ & $0.085\pm0.001$ & $2.091\pm0.008$ \\
        & AC-MTM    & $0.836\pm0.000$ & $0.079\pm0.001$ & $1.777\pm0.054$ \\
        & p-IDM     & $0.838\pm0.000$ & $0.071\pm0.002$ & $2.195\pm0.039$ \\
        & \method{} & $0.839\pm0.000$ & $0.068\pm0.004$ & $1.880\pm0.034$ \\
        \midrule
        \multirow{5}{*}{Reacher}
        & LeWM      & $-0.002\pm0.000$ & $0.191\pm0.000$ & $1.363\pm0.013$ \\
        & SMWM      & $-0.002\pm0.000$ & $0.192\pm0.000$ & $1.389\pm0.014$ \\
        & AC-MTM    & $-0.002\pm0.000$ & $0.192\pm0.000$ & $1.399\pm0.018$ \\
        & p-IDM     & $-0.001\pm0.000$ & $0.193\pm0.000$ & $1.403\pm0.004$ \\
        & \method{} & $-0.001\pm0.000$ & $0.193\pm0.000$ & $1.421\pm0.011$ \\
        \midrule
        \multirow{5}{*}{Cube}
        & LeWM      & $0.735\pm0.001$ & $0.123\pm0.001$ & $3.582\pm0.049$ \\
        & SMWM      & $0.740\pm0.000$ & $0.188\pm0.000$ & $5.022\pm0.027$ \\
        & AC-MTM    & $0.746\pm0.000$ & $0.181\pm0.000$ & $5.227\pm0.038$ \\
        & p-IDM     & $0.751\pm0.000$ & $0.145\pm0.001$ & $4.624\pm0.064$ \\
        & \method{} & $0.753\pm0.000$ & $0.120\pm0.001$ & $4.375\pm0.052$ \\
        \midrule
        \multirow{5}{*}{Finger}
        & LeWM      & $0.348\pm0.001$ & $0.219\pm0.001$ & $1.737\pm0.013$ \\
        & SMWM      & $0.359\pm0.001$ & $0.232\pm0.001$ & $1.850\pm0.009$ \\
        & AC-MTM    & $0.359\pm0.000$ & $0.231\pm0.001$ & $1.857\pm0.008$ \\
        & p-IDM     & $0.360\pm0.000$ & $0.229\pm0.001$ & $1.838\pm0.002$ \\
        & \method{} & $0.356\pm0.001$ & $0.230\pm0.001$ & $1.871\pm0.008$ \\
        \midrule
        \multirow{5}{*}{Ball-in-Cup}
        & LeWM      & $0.818\pm0.000$ & $0.049\pm0.000$ & $1.939\pm0.021$ \\
        & SMWM      & $0.813\pm0.001$ & $0.056\pm0.001$ & $1.748\pm0.095$ \\
        & AC-MTM    & $0.814\pm0.001$ & $0.056\pm0.001$ & $1.744\pm0.029$ \\
        & p-IDM     & $0.813\pm0.000$ & $0.054\pm0.001$ & $1.695\pm0.008$ \\
        & \method{} & $0.810\pm0.000$ & $0.052\pm0.001$ & $1.838\pm0.011$ \\
        \bottomrule
    \end{tabular}
\end{table}

\paragraph{History predictability and next-latent gains.}
History alone explains 74--84\% of action variance on PushT, Cube, and Ball-in-Cup, roughly 35\% on Finger, and virtually none on Reacher (consistent with uniform random actions; \Cref{tab:inverse_head_diagnostics}). This confirms the practical relevance of \Cref{prop:idm}: in realistic offline datasets, most action variance is already predictable before seeing $Z'$. Even so, observing $Z'$ improves both regression and likelihood across every environment and encoder---for \method{}, cutting the remaining history-only MSE by about 27\% on Ball-in-Cup, 42\% on PushT, and 49\% on Cube.

\paragraph{Comparing regression and likelihood across encoders.}
Regression and likelihood gains capture distinct aspects of next-latent predictability. On PushT and Cube, \method{} achieves higher likelihood gains than LeWM while maintaining similar normalized regression gains, whereas SMWM and AC-MTM raise both metrics. On Ball-in-Cup, LeWM obtains the highest likelihood gain, while differences on Finger and Reacher are minor. (Because each encoder alters both $H$ and $Z'$, matching regression gains across encoders do not imply identical conditional means; \Cref{app:direction_distance_example} isolates mean-preserving information cleanly.)

\paragraph{Relation to planning.}
Neither regression nor likelihood gains reliably predict downstream planning success. SMWM and AC-MTM achieve larger regression gains than \method{} on PushT, Cube, and Ball-in-Cup, yet \method{} beats both on PushT and Ball-in-Cup and beats SMWM on Cube (while closely matching AC-MTM on Cube; \Cref{tab:planning_mppi_cem}). Similarly, higher likelihood gains for SMWM (PushT, Cube) and LeWM (Ball-in-Cup) do not translate into higher planning success, and on PushT the higher regression gains of SMWM and AC-MTM coincide with degraded block-orientation readouts (\Cref{tab:representation_readout}). The same pattern holds for p-IDM: it achieves the highest mean likelihood gain on PushT and larger regression and likelihood gains than \method{} on Cube, yet yields lower planner-averaged success than \method{} in both environments (\Cref{tab:smwm_history}). On Reacher and Finger, p-IDM's predictive gains are close to those of the other encoders, whereas on Ball-in-Cup it records the lowest mean likelihood gain among all encoders alongside weak planning performance. Echoing our candidate-source ablation (\Cref{tab:negsrc_score_replacement,tab:negsrc_planners}), these findings show that raw inverse predictability is not a proxy for planning quality; auxiliary supervision should be judged by the goal-relevant state distinctions it preserves.

\section{Qualitative Examples}

We show matched planning rollouts starting from identical initial and goal states across methods, highlighting both successful \method{} trajectories and a remaining failure mode on Cube. Each frame is scored under the criteria in \Cref{app:env_dataset}: green borders mark goal satisfaction, and red borders on final frames mark failure at step 50. Under the any-time evaluation rule, a rollout can succeed earlier even if its final frame ends outside tolerance.

\begin{figure}[t]
    \centering
    \includegraphics[width=\linewidth]{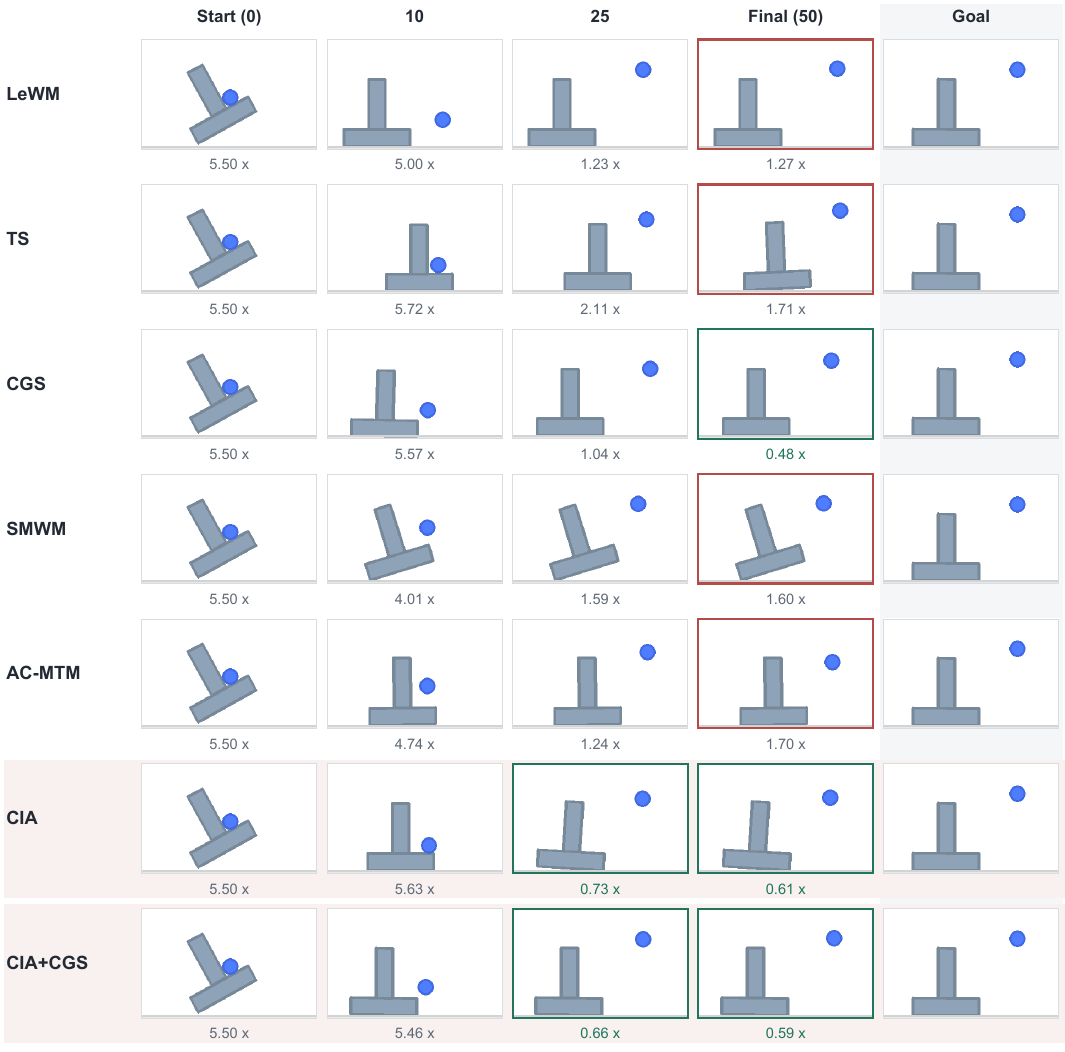}
    \caption{\textbf{PushT: reaching the pusher and block goals.} Matched CEM rollouts at steps 0, 10, 25, and 50, with the goal at right. \method{}, CGS, and \method{}+CGS satisfy the final goal. Values show the maximum of the concatenated pusher--block Euclidean position error normalized by 20 pixels and the wrapped block-angle error normalized by $20^\circ$; values below one indicate success.}
\end{figure}

\begin{figure}[t]
    \centering
    \includegraphics[width=\linewidth]{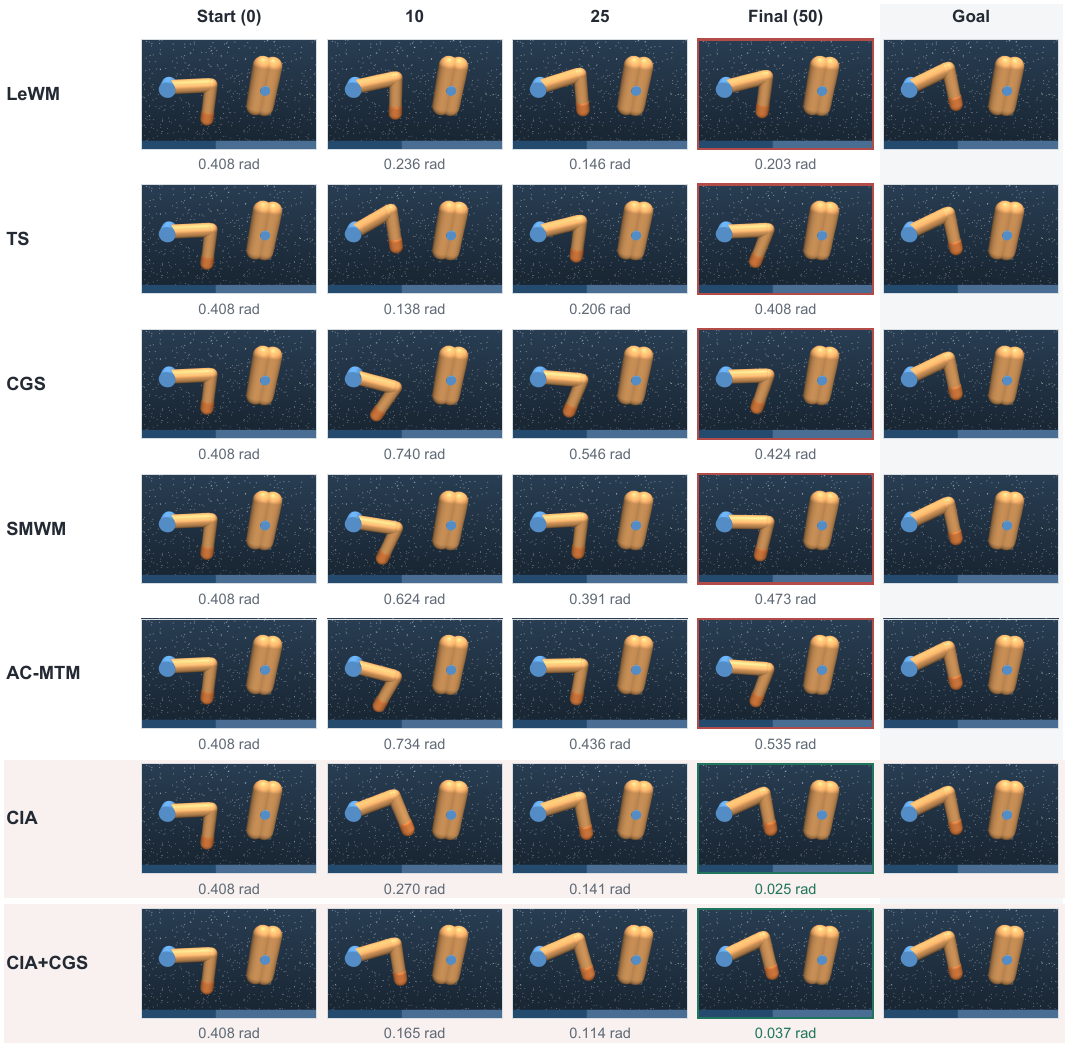}
    \caption{\textbf{Finger: completing the joint configuration.} Matched CEM rollouts at steps 0, 10, 25, and 50, with the goal at right. \method{} and \method{}+CGS satisfy the final goal. Values show the maximum absolute error across the two finger joints and spinner hinge, without angle wrapping; success requires an error below 0.05 radians.}
\end{figure}

\begin{figure}[t]
    \centering
    \includegraphics[width=\linewidth]{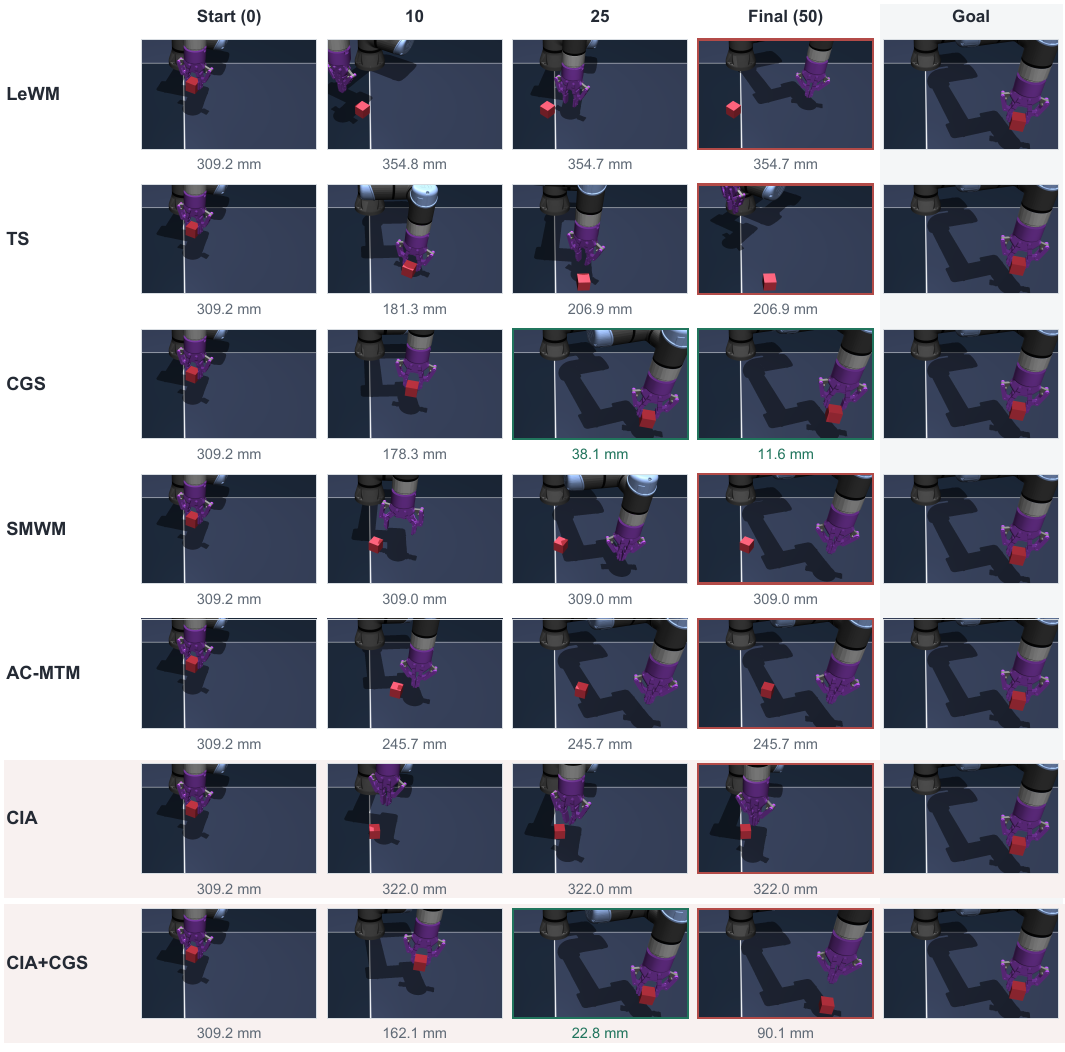}
    \caption{\textbf{Cube: a remaining failure during object transport.} Matched CEM rollouts at steps 0, 10, 25, and 50, with the goal at right. \method{} drops the initially held cube, whereas CGS satisfies the final goal. \method{}+CGS succeeds during execution but finishes outside tolerance. Values show the Euclidean cube-position error; success requires an error below 40 mm.}
\end{figure}

\begin{figure}[t]
    \centering
    \includegraphics[width=\linewidth]{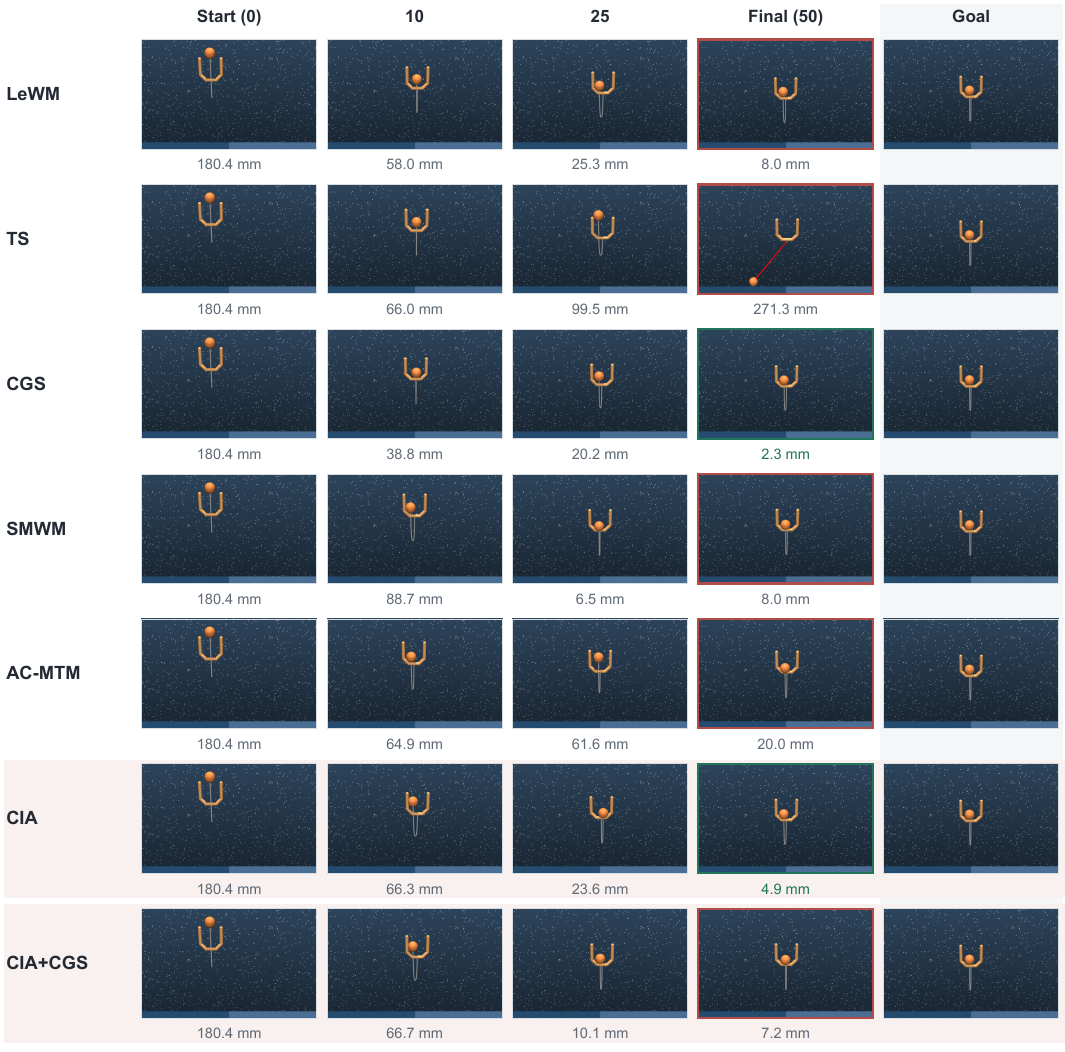}
    \caption{\textbf{Ball-in-Cup: reaching the goal from an outside-cup start.} Matched MPPI rollouts at steps 0, 10, 25, and 50, with the goal at right. \method{} and CGS satisfy the final goal. SMWM, AC-MTM, and \method{}+CGS succeed earlier, possibly between displayed frames, but finish outside tolerance. Values show the maximum absolute error across the cup and ball position coordinates; success requires each error to be below 5 mm.}
\end{figure}

\end{document}